\documentclass[10pt,twocolumn,letterpaper]{article}

\usepackage{iccv}
\usepackage{times}
\usepackage{epsfig}
\usepackage{graphicx}
\usepackage{amsmath}
\usepackage{amssymb}
\usepackage{booktabs}
\usepackage{xcolor}         
\usepackage[accsupp]{axessibility}

\usepackage{pifont}

\usepackage[breaklinks=true,colorlinks,bookmarks=false]{hyperref}

\iccvfinalcopy 

\def\iccvPaperID{****} 

\ificcvfinal\fi

\newcommand{\Fref}[1]{Figure \ref{#1}}

\newcommand{\Tref}[1]{Table \ref{#1}}

\newcommand{\fref}[1]{figure \ref{#1}}

\begin{document}

\title{Improving Diversity in Zero-Shot GAN Adaptation with Semantic Variations}

\author{Seogkyu Jeon$^{1}$\thanks{Work done during his internship at Microsoft Research Asia} \quad Bei Liu$^{2}$ \quad Pilhyeon Lee$^{1}$ \quad Kibeom Hong$^{1,3}$ \quad Jianlong Fu$^{2}$ \quad Hyeran Byun$^{1}$\thanks{Corresponding author}\vspace{0.2cm}\\
$^{1}$Yonsei University\quad\quad $^{2}$Microsoft Research Asia\quad\quad $^{3}$SwatchOn\\
{\tt\small jone9312@yonsei.ac.kr}
}

\maketitle
\ificcvfinal\thispagestyle{empty}\fi

\begin{abstract}
   Training deep generative models usually requires a large amount of data. To alleviate the data collection cost, the task of zero-shot GAN adaptation aims to reuse well-trained generators to synthesize images of an unseen target domain without any further training samples. Due to the data absence, the textual description of the target domain and the vision-language models, e.g., CLIP, are utilized to effectively guide the generator. However, with only a single representative text feature instead of real images, the synthesized images gradually lose diversity as the model is optimized, which is also known as \textit{mode collapse}. To tackle the problem, we propose a novel method to find semantic variations of the target text in the CLIP space. Specifically, we explore diverse semantic variations based on the informative text feature of the target domain while regularizing the uncontrolled deviation of the semantic information. With the obtained variations, we design a novel directional moment loss that matches the first and second moments of image and text direction distributions. Moreover, we introduce elastic weight consolidation and a relation consistency loss to effectively preserve valuable content information from the source domain, e.g., appearances. Through extensive experiments, we demonstrate the efficacy of the proposed methods in ensuring sample diversity in various scenarios of zero-shot GAN adaptation. We also conduct ablation studies to validate the effect of each proposed component. Notably, our model achieves a new state-of-the-art on zero-shot GAN adaptation in terms of both diversity and quality.
\end{abstract}

\section{Introduction}
\label{sec:intro}
In recent years, deep generative models, especially generative adversarial networks~(GANs)~\cite{goodfellow2014generative}, have shown dramatic advancements by successfully mimicking the real distribution of images~\cite{brock2018large,karras2019style,dhariwal2021diffusion}.
However, as diagnosed in the literature~\cite{zhao2020differentiable,luvcic2019high}, building powerful generative models requires a huge number of visual samples as well as expensive training costs.
This essentially restricts the applicability of the models to domains where it is prohibitively expensive or even infeasible to collect sufficient data, such as medical images or artworks of specific artists.
To alleviate the limitation, researchers give their attention to the task of GAN adaptation, which aims to reuse the representation power of well-trained generators for synthesizing images of a target domain.
To this end, existing works manage to transfer the generation capability of pre-trained GANs to unseen target domains by exploiting a tiny dataset~\cite{wang2018transferring} with only a few visual samples~(\ie, few-shot)~\cite{wang2020minegan, li2020few, liu2020towards, noguchi2019image, robb2020few, ojha2021few}, or even no data at all~(\ie, zero-shot)~\cite{gal2021stylegan}.
This paper focuses on the zero-shot setting, where 
a generative model pre-trained on a source dataset is supposed to be adapted to an unseen target domain that contains no visual samples for training.

In order to perform adaptation with no accessibility to data of the target domain, the previous work hinges on the powerful vision-language model, \ie, CLIP~\cite{radford2021learning}, which learns the shared latent space between vision and text modalities.
Specifically, StyleGAN-NADA~\cite{gal2021stylegan} embeds two textual prompts respectively describing the source and target domains into the CLIP space and derives the difference vector between them.
Considering the difference vector to be the guiding direction, the generated images gradually step toward the target domain.
Eventually, the generator is able to synthesize visually plausible images of the target domain even without seeing any samples of the domain.


\begin{figure*}[!t]
  \centering
  \includegraphics[clip=true,width=0.97\linewidth]{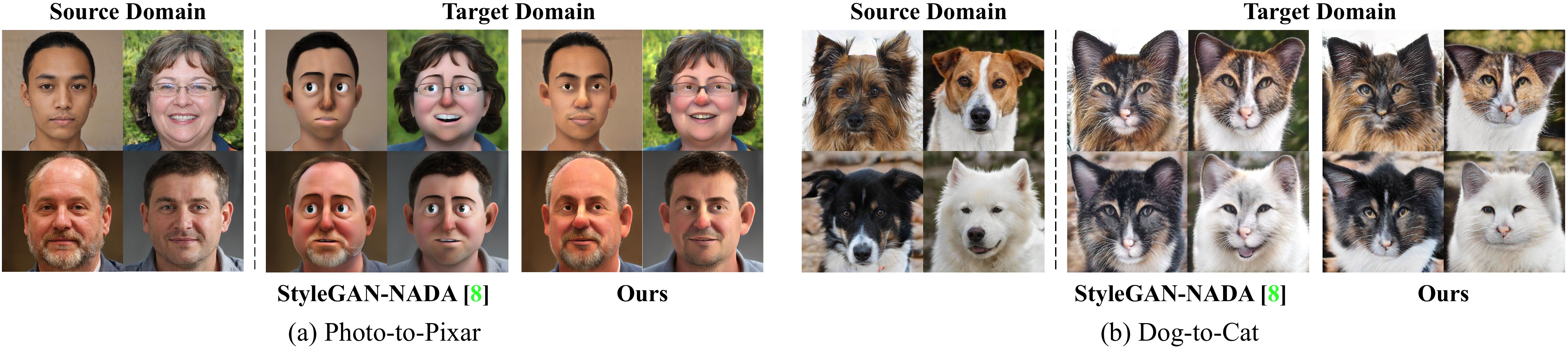}
  \caption{Illustration of our motivation. For two adaptation scenarios, \ie, ``Photo-to-Pixar'' and ``Dog-to-Cat'', we present source domain images and the corresponding generated images of the target domain by StyleGAN-NADA~\cite{gal2021stylegan} and ours.}
  \label{fig:teaser}
\end{figure*}

However, the adapted model with only a single guiding direction suffers from \textit{mode collapse}. That is, the generated target samples share the same characteristics without distinction. For instance, the generated faces under the ``Photo-to-Pixar'' scenario have exactly the same attributes, \eg, emotional expression, slightly opened mouth, dark hair~(\fref{fig:teaser} (a) center). In another example, the results under the ``Dog-to-Cat'' scenario exhibit nearly identical cat faces with few differences (\fref{fig:teaser} (b) center). These problems come from the one-to-one mechanism of the CLIP text encoder; given a single guiding direction, the adapted generator is unable to handle the diversity of the target domain.
Intuitively, the target textual prompt provides the most general information about the target domain while there are an infinitely large number of semantic variations underneath.
For instance, the target domain ``Cat'' implicitly covers a variety of species with diverse characteristics such as a smiling Sphynx and a brown Scottish Fold.
Hence, relying solely on a single target description fails to exploit the inherent semantic variations, and it is crucial to model a one-to-many relation for enhancing sample diversity after adaptation.


In this paper, we explore semantic variations of the given text prompt of the target domain with a novel two-stage framework in order to alleviate the mode collapse problem. In the first stage, to discover the variations, we impose a set of learnable perturbations on the target text embedding in the CLIP space.
They are encouraged to be orthogonal to each other for redundancy reduction yet not to disturb the original semantics of the target domain.
The obtained variations are then used to compute guiding directions.
In the second stage, to improve sample diversity using multiple guidances, we introduce a novel \textit{directional moment loss}.
It effectively aligns image-updating directions with the guidances by matching their first and second moments.

In addition, we propose a \textit{relation consistency loss} to better sustain the knowledge of the generator learned from the source domain. Ideally, the relation between two generated images should remain the same during adaptation to ensure consistency of semantic information.
From this motivation, the relation consistency loss is designed to minimize the distribution gap between the inter-image relations of the source and target domains.
Further, we employ the elastic weight consolidation~\cite{kirkpatrick2017overcoming} to suppress excessive changes of important parameters of the generator during adaptation.
This prevents our model from losing the strong content representability of the generator during the adaptation, thereby preserving the original content information.

Equipped with the proposed components, our model is able to generate images with diverse semantic variations of the target domain, while successfully preserving the original semantic information of the source domain.
The superiority of our method over the previous work is clearly showcased in \Fref{fig:teaser}.
Through extensive experiments on various adaptation scenarios, we demonstrate the effectiveness of each component of our model.
Moreover, our model achieves a new state-of-the-art on zero-shot GAN adaptation in terms of quality as well as diversity.

\section{Related Works}
\label{sec:related_work}

\begin{figure*}[t]
  \centering
  \includegraphics[clip=true,width=0.98\linewidth]{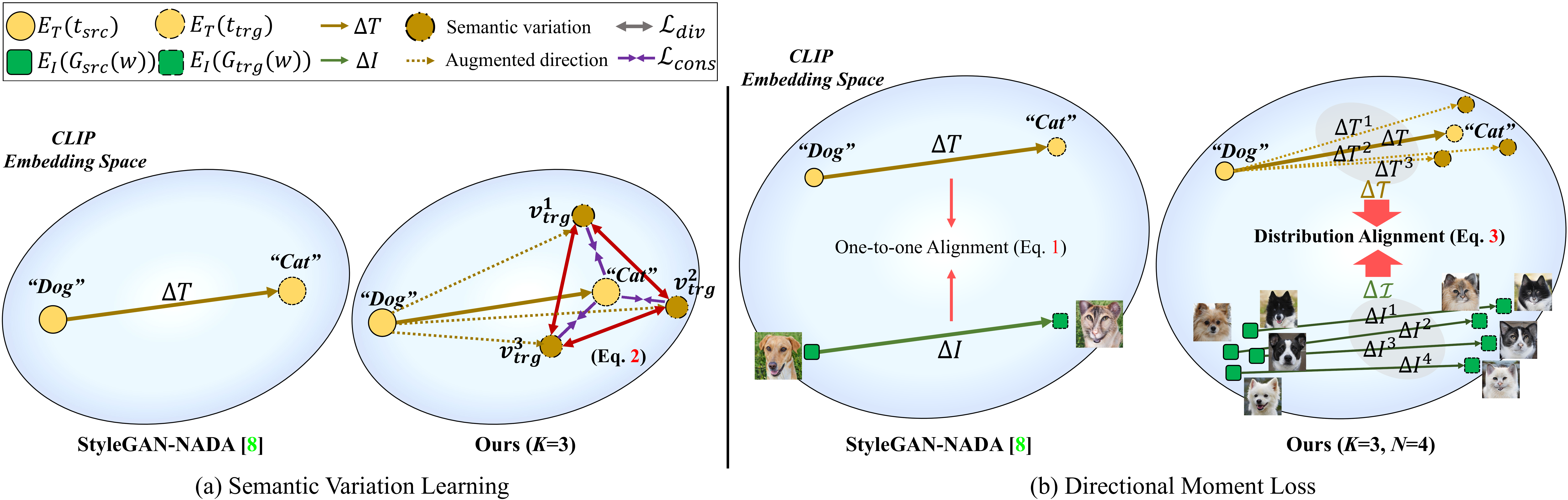}
  \caption{Illustration of the proposed methods. (left) semantic variation learning in CLIP space (right) the directional moment loss. Here $K$ denotes the number of augmented semantic variations of the target text $v^{i}_{trg}$, while $N$ is the batch size of generated samples.}
  \label{fig:architecture}
\end{figure*}
\paragraph{Few-shot GAN adaptation.}
In the last decade, research on deep generative models has achieved remarkable advances and they are now capable of almost completely mimicking the distributions of real images~\cite{brock2018large, karras2019style, karras2020analyzing, karras2021alias}. However, on the dark side, they require an excessively large amount of real images for effective and stable training from scratch. Constructing a large-scale well-refined training dataset is excessively costly and laborious, and even unavailable in some domains, \eg, artworks. To this end, several studies~\cite{karras2020training, zhao2020differentiable, tseng2021regularizing, tran2020towards, zhang2019consistency, zhao2021improved} are proposed to accomplish data-efficient training with a small number of training samples provided (\eg, $10^3$ to $10^4$).


Despite their achievements, the studies still struggle in a more restrictive setting where only a few samples less than 10 are accessible. Due to the formidable data scarcity, the generator is prone to overfitting, \ie, memorizing only some training samples, thereby losing diversity and falling into mode collapse.
To tackle the problem, the task of few-shot GAN adaptation~\cite{mo2020freeze,wang2020minegan,li2020few,liu2020towards,noguchi2019image,robb2020few,wang2018transferring,ojha2021few, zhang2022generalized, xiao2022few, zhao2022closer, zhu2021mind, kim2022dynagan} arises to adapt well-trained generative models to the target domain. As the knowledge of the target domain is largely limited, pre-trained GANs are generally leveraged to distill the diverse content information learned from the large-scale dataset. MineGAN~\cite{wang2020minegan} introduces the mining network to identify beneficial knowledge for the target domain generation.
Meanwhile, Li~et~al.~\cite{li2020few} preserve the weights of the generative models with elastic weights consolidation~\cite{kirkpatrick2017overcoming} based on Fisher information. Ojha~et~al.~\cite{ojha2021few} propose a GAN adaptation framework with a cross-domain correspondence loss and a relaxed discriminator. RSSA~\cite{xiao2022few} proposes a relaxed spatial consistency method that encourages the generator maintain the self-correlation and the inter-sample spatial correlation. DCL~\cite{zhao2022closer} proposes a contrastive learning framework to enhance visual quality and ameliorate diversity degradation.


\paragraph{Zero-shot GAN adaptation.}


Recently, CLIP~\cite{radford2021learning} has brought a significant impact on the computer vision fields, showing impressive performance as well as robustness in the zero-shot classification task. Thanks to its powerful cross-modal representation, recent research~\cite{patashnik2021styleclip, gabbay2021image, cheng2021data, narasimhan2021clip, zabari2021semantic, gal2021stylegan, patashnik2021styleclip, kwon2021clipstyler, kim2021diffusionclip} have actively attempted to exploit the pre-trained CLIP for generative tasks.

Among others, bringing the power of CLIP to GAN, StyleGAN-NADA~\cite{gal2021stylegan} enables GAN adaptation only with the textual descriptions of the target domain without any training images, \ie, zero-shot GAN adaptation.
Using the guiding direction obtained from the textual domain descriptions, it adapts the generator so that the generated images move in accordance with the guidance in the CLIP space.
As a result, it can synthesize samples of the target domain by relying on only textual descriptions, not image samples.
Unlike the previous latent manipulation method~\cite{patashnik2021styleclip} where available modifications are constrained in the domain of the pre-trained generator, zero-shot GAN adaptation can perform out-of-domain manipulation by directly optimizing the generator parameters.


However, as shown in \Fref{fig:teaser}, StyleGAN-NADA fails to capture diverse semantic variations of the target text, resulting in the mode collapse of generated samples. We conjecture that this is due to its adaptation process that relies heavily on a single target domain description. To handle this problem, we propose to discover semantic variations of the target text in CLIP space, which enables generating diverse samples of the target domain in the zero-shot setting.

\section{Proposed Methods}
\label{sec:propsed_method}

\subsection{Baseline}
Our baseline for text-driven GAN adaptation is similar to StyleGAN-NADA~\cite{gal2021stylegan} except that we do not use complex layer selection. Its architecture basically follows StyleGAN2~\cite{karras2020analyzing} which consists of a mapping network and a generator $G_{src}$. The mapping network is trained to embed a latent code from the prior distribution into the disentangled latent space $\mathcal{W}$. The generator $G_{src}$ takes the converted latent code $w \in \mathbb{R}^{B \times D_{w}}$ as input to generate RGB images $G_{src}(w) \in \mathbb{R}^{N \times 3 \times H \times W}$ of the training domain. Here $D_{w}$ is the dimension of the latent space, $N$ denotes the batch size, and $(H, W)$ indicates the size of generated images.

The main training objective is to adapt the pre-trained generator~$G_{src}$ on a source domain (\eg, cat) to synthesize the images of a target domain (\eg, dog) using the descriptions of the source and target domains, \ie, $t_{src}$ and $t_{trg}$, as the text prompt. Note that $G_{src}$ is sufficiently optimized to generate realistic samples of the source domain. The target domain generator $G_{trg}$ is initialized by the parameters of $G_{src}$ and optimized during training, whereas $G_{src}$ and the mapping network remain frozen. By doing so, we can generate source samples $G_{src}(w)$ and target samples $G_{trg}(w)$ from the same latent code $w$ to estimate the image-level relation between two domains. To convey the learned knowledge of $G_{src}$ to $G_{trg}$, the directional loss $\mathcal{L}_{dir}$ is designed to align the direction between source and target images with the text direction in the CLIP embedding space by maximizing their cosine similarities as follows.
\begin{equation}
\label{equ:directional_loss}
\begin{split}
    \mathcal{L}_{dir} &= \frac{1}{N}\sum_{n=1}^{N} \Big[ 1 - \frac{\Delta{I^n}\cdot\Delta{T}}{\left\|\Delta{I^n}\right\|\left\|\Delta{T}\right\|}\Big],
    \\
    \text{where}~~\Delta{T} &= E_{T}(t_{trg}) - E_{T}(t_{src})  \\
    \text{and}~~\Delta{I}^{n} &= E_{I}(G_{trg}(w^{n})) - E_{I}(G_{src}(w^{n})). \\
\end{split}
\end{equation}
Here $N$ is the mini-batch size, while $E_{T}$ and $E_{I}$ respectively denote the text encoder and the image encoder of the pre-trained CLIP~\cite{radford2021learning} that share the same embedding space with its dimension of $D$.
It is worth noting that the directional loss encourages \textit{every} image sample to be updated in the same direction with the text guidance, \ie, one-to-one alignment (see \Fref{fig:architecture} (b) left).



\subsection{Motivation}
As the CLIP text encoder is in nature deterministic, the directional loss $\mathcal{L}_{dir}$ is computed with only a single direction toward the representative feature of the target domain, \ie, $\Delta T$.
Consequently, all generated image samples are updated in the same direction and encouraged to share typical characteristics, while gradually diminishing the diversity during the adaptation. Utilizing some textual templates to decorate the target text can be deemed an intuitive solution, but manually defining templates for each specific target domain is heuristic and less generalizable. Instead, we propose to augment the features by exploring \textit{semantic variations}, in order to alleviate the mode collapse. Here, the semantic variations denote feature vectors that are semantically consistent with the target text while being capable of expressing diverse characteristics.

The overall pipeline of our method consists of two stages. In the first stage, we search for diverse semantic variations to augment the target text feature $E_{T}(t_{trg})$ while preserving its original information. In the second stage, we guide the target generator $G_{trg}$ with the augmented text directions while maintaining sample diversity.

\subsection{Semantic Variation Learning}

To find semantic variations in the CLIP space, we first prepare $K$ learnable vectors $\{z^{i}\}_{i=1}^{K}$, where $K$ denotes the number of variations and the vectors share the same space with the target text feature, \ie, $z^i \in \mathbb{R}^{D}$.
Thereafter, each vector serves as an additive perturbation on the target text feature and learns a useful semantic variation that can enhance the diversity but does not disturb the original semantics of the target domain text.
Concretely, the vectors are optimized with two loss functions as follows.
\begin{equation}
\label{equ:semantic_variation_optimization}
\begin{split}
    \mathcal{L}_{cons} &= \frac{1}{K}\sum^{K}_{i=1}\Big[ 1 - \frac{E_T(t_{trg})\cdot {v}^{i}_{trg}}{\left\|E_T(t_{trg})\right\|\left\| {v}^{i}_{trg}\right\|} \Big], \\
    \mathcal{L}_{div} &= \binom{K}{2}^{-1}\sum^{K-1}_{i=1}\sum^{K}_{j=i+1}\left|\frac{z^i \cdot z^j}{\left\|z^i\right\|\left\|z^j\right\|}\right|, \\
\end{split}
\end{equation}
where ${v}^{i}_{trg}= E_{T}(t_{trg}) + \epsilon\frac{z^{i}}{\left\|z^{i}\right\|}$ denotes the perturbed text feature, \ie, the semantic variation, and $\epsilon$ is a hyperparameter determining the perturbation strength. $\mathcal{L}_{cons}$ is a semantic consistency loss that prevents unintended deviation of the original semantic by regularizing the cosine distance between the original text feature $E_{T}(t_{trg})$ and its semantic variations $v^{i}_{trg}$.
On the other hand, $\mathcal{L}_{div}$ is a semantic diversity loss that prevents the perturbation vectors from learning redundant information by encouraging orthogonality for all combinations of $z^*$.

To summarize, we search the semantic variations $\{v^i_{trg}\}_{i=1}^{K}$ by optimizing $\{z^{i}\}_{i=1}^{K}$ with the weighted sum of the losses $\mathcal{L}_{S1} = \mathcal{L}_{cons} + \lambda_{div} \mathcal{L}_{div}$, where $\lambda_{div}$ is a weighting factor.
Our semantic variation learning is depicted in \Fref{fig:architecture} (a).

\subsection{Directional Moment Loss}
After searching the semantic variations $\{v^{i}_{trg}\}_{i=1}^{K}$, we utilize them as a kind of augmentation to guide $G_{trg}$ with multiple directions between the source and target texts. To encourage $G_{trg}$ to learn the diversity from a single target text and its semantic variations, we propose a novel \textit{directional moment loss}. In specific, we first compute the text direction from the source text feature $E_T(t_{src})$ to each semantic variation $v^{i}_{trg}$ as $\Delta{T^{i}} = v^{i}_{trg} - E_T(t_{src})$. Then we compose the text direction set with the original direction $\Delta{T}$ and perturbed ones $\Delta{T}^{*}$ as $\Delta{\mathcal{T}} = [\Delta{T}; \Delta{T^{1}}; \dots; \Delta{T^{K}}]^\top \in \mathbb{R}^{(K+1) \times D}$, where $D$ denotes the channel dimension.
Meanwhile, the image direction set can be obtained by composing the image directions within the batch: $\Delta{\mathcal{I}} = [\Delta{I^{1}}; \Delta{I^{2}}; \dots; \Delta{I^{N}}]^\top \in \mathbb{R}^{N \times D}$.
We design a directional moment loss to minimize the distances between the image and the text direction sets by matching their first and second moments.
Specifically, we align the mean of the image direction set $\mu_{\Delta{\mathcal{I}}} = \frac{1}{N}\sum^{N}_{n=1}\Delta{\mathcal{I}_n}$ with the mean of the text direction set $\mu_{\Delta{\mathcal{T}}} = \frac{1}{K+1}\sum_{i=1}^{K+1}\Delta{\mathcal{T}_i}$, while matching the covariance of the image direction set $\Sigma_{\Delta{\mathcal{I}}} = {\Delta{\mathcal{I}}^\top\Delta{\mathcal{I}}}$ with that of the text direction set $\Sigma_{\Delta{\mathcal{T}}}=\Delta{\mathcal{T}}^\top\Delta{\mathcal{T}}$. The directional moment loss is defined as:
\begin{equation}
\label{equ:directional_moment_loss}
    \mathcal{L}_{dm} = d_{1}(\mu_{\Delta{I}}, \mu_{\Delta{\mathcal{T}}}) + \lambda_{cov} d_{2}(\Sigma_{\Delta{I}},\Sigma_{\Delta{\mathcal{T}}}),
\end{equation}
where $\lambda_{cov}$ is a balancing weighting factor. We instantiate $d_{1}(\cdot, \cdot)$ with the cosine distance and $d_{2}(\cdot, \cdot)$ with the euclidean distance. Note that by adding the second term, we can prevent the image directions from being collapsed into a single direction, thus ensuring the sample diversity.
A conceptual illustration of our directional moment loss is provided in Figure~\ref{fig:architecture} (b) right.

\subsection{Source Knowledge Preservation}
To enhance the realism after the adaptation, it is important to preserve valuable content information such as appearances learned from the source domain. For this purpose, StyleGAN-NADA~\cite{gal2021stylegan} utilizes the layer selection strategy to estimate the importance of each layer and select the top-$k$ important layers to be updated while freezing the rest. However, the number of layers, \ie, $k$, needs to be tuned for each adaptation scenario, which is cumbersome. Moreover, the frozen layers can also contain valuable information for synthesizing realistic content. In this point of view, we propose to constrain each layer in accordance with its importance for the original task, \ie, source domain generation. To this end, we employ the elastic weight consolidation~(EWC)~\cite{kirkpatrick2017overcoming} to penalize drastic modification of model parameters. The EWC regularization loss is formulated as:
\begin{equation}
\label{equ:ewc_loss}
    \mathcal{L}_{EWC} = \sum^{L}_{l=1}F^{l}(\theta^{l}_{trg}-\theta^{l}_{src})^{2},
\end{equation}
where $l$ is the layer index, while $\theta$ is the trainable parameters of the generator.
The fisher information~\cite{pascanu2013revisiting} is estimated as $F = \mathbb{E}\left [-\frac{\partial^2}{\partial\theta^2_{src}}sim(E_{I}(G_{src}(w)), E_{T}(t_{src})) \right ]$, where $sim(\cdot, \cdot)$ is the cosine similarity between the source text and the generated source samples in the CLIP embedding space.

For more diversity, we design a relation consistency loss that encourages the generator to maintain the semantic relation between images before and after adaptation. Specifically, we extract image features from the generated samples of source and target domain, \ie, $x_{src} = E_{I}(G_{src}(w)) \in \mathbb{R}^{N \times D}$. With the extracted features, we estimate the inter-relation of samples by calculating their dot product similarities $M_{src} = x_{src} \cdot x_{src}^\top \in \mathbb{R}^{N \times N}$, which denotes the inter-sample relation matrix of source samples. The inter-sample relation matrix for target samples $M_{trg}$ can be obtained in a similar way. We apply the row-wise softmax function to $M_{src}$ and $M_{trg}$, and then minimize the KL divergence to make the target relation similar to the source one.
The relation consistency loss is defined as:
\begin{equation}
\label{equ:relation_consistency_loss}
    \mathcal{L}_{rel} = KL(\text{Softmax}(M_{src}), \text{Softmax}(M_{trg})).
\end{equation}







The overall training objective of the target generator $G_{trg}$ in the second stage, \ie, the adaptation stage, is the weighted sum of the loss functions with two balancing factors $\lambda_{EWC}$ and $\lambda_{rel}$ as follows.
\begin{equation}
\label{equ:overall_loss}
    \mathcal{L}_{S2} = \mathcal{L}_{dm} + \lambda_{EWC}\mathcal{L}_{EWC} + \lambda_{rel}\mathcal{L}_{rel}.
\end{equation}

\section{Experiments}
\label{sec:experiments}
\subsection{Implementation Details}
Following the setting of the previous work~\cite{gal2021stylegan}, we implement our method based on StyleGANv2~\cite{karras2020analyzing} pre-trained on FFHQ~\cite{karras2019style}, AFHQ-Dog, and AFHQ-Cat~\cite{choi2020stargan} datasets. The text descriptions of source domains for FFHQ, AFHQ-Dog, and AFHQ-Cat are set to ``Photo'', ``Dog'', and ``Cat'', respectively.
In stage 1, the semantic variations $\{z^i\}_{i=1}^{K}$ are optimized with $\mathcal{L}_{S1}$ during 2,000 iterations.
We set the perturbation strength $\epsilon$ to the $l2$-norm of the $E_T(t_{trg})$, which is empirically shown to be simple yet effective. The number of semantic variations $K$ is fixed to 6 during evaluation. The balancing weight $\lambda_{div}$ is set to 1.
In stage 2, the generator $G_{trg}$ is trained with the batch size $N$ of 4.
For the vision-language model, we employ the pre-trained CLIP~\cite{radford2021learning} with the image encoder of ViT-B/32.
The weighting factors $\lambda_{cov}$, $\lambda_{EWC}$, and $\lambda_{rel}$ are set to $10^{3}$, $10^{7}$, and $10^{2}$ regarding their loss scales. We utilize the Adam~\cite{kingma2014adam} optimizer with the learning rate of 0.002 with betas of $(0, 0.99)$ for variations $\{z^i\}_{i=1}^{K}$ as well as the generator $G_{trg}$. We conduct all the experiments on a single RTX 2080Ti GPU.


\subsection{Quantitative Results}


\paragraph{Diversity comparison.} Ideally, the generator after adaptation should synthesize the samples of the target domain well while preserving the semantic variations learned from the source domain.
To evaluate how well the semantic variations are maintained, we compute the \textit{intra-cluster pairwise LPIPS distance} that directly measures the diversity of generated samples following the existing work~\cite{ojha2021few}.
This metric is originally designed for the few-shot setting, where the individual training images are considered to be cluster centroids and the generated samples are clustered using LPIPS~\cite{zhang2018unreasonable}.
Thereafter, the average LPIPS distance within the cluster is estimated to represent the generated sample diversity.
Following StyleGAN-NADA~\cite{gal2021stylegan}, we adapt the metric to the zero-shot setting by building a total of $k$ clusters using $k$-medoids clustering~\cite{kaufman1990partitioning}.
For evaluation, we generate 1,000 samples of the target domain for evaluation and compare our method with the state-of-the-art zero-shot method, \ie, StyleGAN-NADA~\cite{gal2021stylegan}, and the few-shot method, \ie, Ojha~et~al.~\cite{ojha2021few}. We set $k$ to 10 for a fair comparison with Ojha~et~al.~\cite{ojha2021few} that utilizes 10-shot data.

\begin{table}[t]
\caption{Quantitative results under the ``Dog-to-Cat'' scenario on AFHQ datasets~\cite{choi2020stargan}.}
  \label{tab:intra_cluster_lpips}
  \centering
\resizebox{\linewidth}{!}{
\begin{tabular}{l|cc}
\toprule
Methods & LPIPS (Avg.) ($\uparrow$) & LPIPS (All) ($\uparrow$)\\
\midrule\midrule

Ojha et al. (10-shot)~\cite{ojha2021few} & 0.575$_{\pm\text{0.019}}$ & 0.575$_{\pm\text{0.046}}$\\

StyleGAN-NADA~\cite{gal2021stylegan} & 0.460$_{\pm\text{0.010}}$ & 0.462$_{\pm\text{0.063}}$\\
StyleGAN-NADA~\cite{gal2021stylegan}~+~$\mathcal{L}_{EWC}$ & 0.480$_{\pm\text{0.006}}$ & 0.480$_{\pm\text{0.064}}$\\
\midrule
Baseline ($\mathcal{L}_{dir}$) & 0.402$_{\pm\text{0.008}}$ & 0.405$_{\pm\text{0.057}}$\\
~~+~replacing $\mathcal{L}_{dir}$ with $\mathcal{L}_{dm}$ & 0.464$_{\pm\text{0.013}}$ & 0.470$_{\pm\text{0.064}}$\\
~~+~$\mathcal{L}_{EWC}$ & 0.493$_{\pm\text{0.015}}$  & 0.497$_{\pm\text{0.067}}$\\
~~+~$\mathcal{L}_{rel}$ (Ours) & \textbf{0.507}$_{\pm\text{0.016}}$  & \textbf{0.512}$_{\pm\text{0.072}}$\\
\bottomrule
\end{tabular}
}
\end{table}


\Tref{tab:intra_cluster_lpips} presents the comparison results. The StyleGAN-NADA (second row) denotes the full model equipped with the directional loss $\mathcal{L}_{dir}$ and the layer selection strategy. Noticeably, our method records the average intra-cluster LPIPS score of $0.507$, significantly outperforming the existing state-of-the-art zero-shot method, StyleGAN-NADA ($0.460$).
Moreover, our model effectively bridges the gap with the 10-shot method~\cite{ojha2021few} even without using any training examples of the target domain.
These results clearly demonstrate the effectiveness of our method in enhancing the diversity of the generated target domain samples.

To better understand where the improvements come from, we break down our method and analyze the effect of each component.
Initially, our baseline model equipped with the directional loss $\mathcal{L}_{dir}$~(Eq.~\ref{equ:directional_loss}) shows very poor diversity performance with the average LPIPS of $0.402$, indicating that relying on the single target description hinders the model from generating diverse images of the target domain.
When augmenting the semantic variations and replacing $\mathcal{L}_{dir}$ with our directional moment loss~$\mathcal{L}_{dm}$~(Eq.~\ref{equ:directional_moment_loss}) under our two-stage framework, the diversity of the generated samples is remarkably improved ($0.402 \rightarrow 0.464$), which verifies the importance of modeling the one-to-many relation of the target domain description.
On the other hand, the proposed elastic weight consolidation loss~$\mathcal{L}_{EWC}$~(Eq.~\ref{equ:ewc_loss}) and the relation consistency loss~$\mathcal{L}_{rel}$~(Eq.~\ref{equ:relation_consistency_loss}) respectively bring additional gains of about $0.029$ and $0.014$, resulting in the final average LPIPS score of $0.507$. This demonstrates the complementarity of the proposed components. To further see the advantage of our elastic weight consolidation loss over the layer selection strategy of StyleGAN-NADA~\cite{gal2021stylegan}, we try applying $\mathcal{L}_{EWC}$ to StyleGAN-NADA instead of the layer selection.
As a result, the average LPIPS score is boosted by $0.02$, which indicates that suppressing the dramatic changes of important parameters using our $\mathcal{L}_{EWC}$ is more effective than manually selecting the number of layers to be frozen. On the other hand, our full model still surpasses the `StyleGAN-NADA $+ \mathcal{L}_{EWC}$' variant to a large extent, which manifests the importance of exploring semantic variations for sample diversity.

\begin{figure}[t]
  \centering
  \includegraphics[width=1.0\columnwidth]{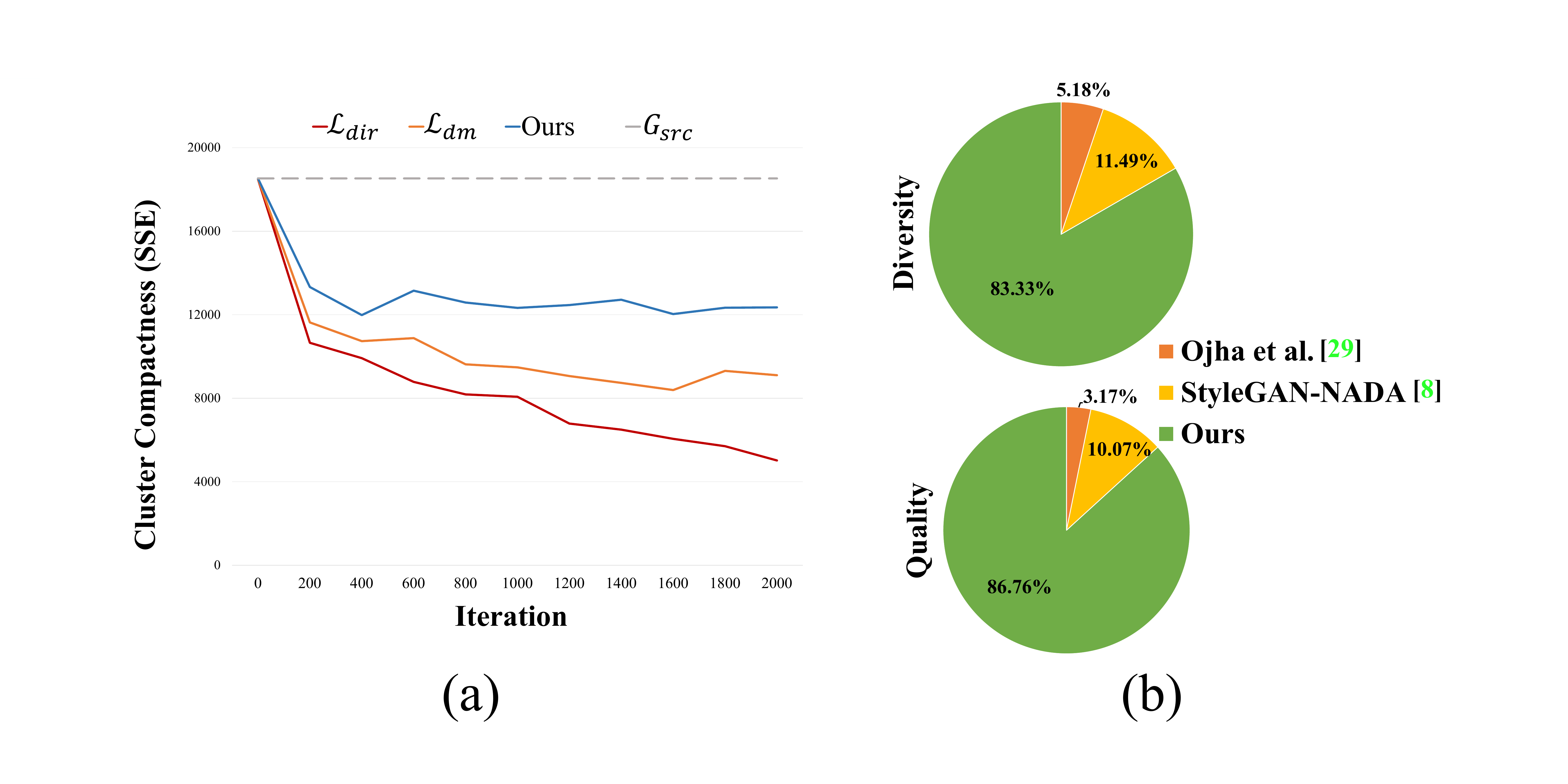}
  \caption{(a) Quantitative comparison of the cluster compactness for diversity evaluation. Higher SSE indicates higher diversity. (b) User study results on the ``Cat-to-Dog'' scenario. Each participant respond to two questions of preference: 1) quality and 2) diversity of generated images.}
  \label{fig:diversity}
\end{figure}

In addition, we also evaluate the diversity with the cluster compactness of generated samples from $G_{trg}$. We regard generated samples from $G_{trg}$ as a single cluster and estimate the cluster compactness by calculating the sum of squared errors~(SSE). To be more specific, we extract features $E_{I}(G_{trg}(w))$ from the CLIP image encoder, \ie, ViT-B/32, and then compute the cluster center as the mean of image features. SSE is in turn estimated with the sum of squared errors from each feature to the cluster center. In~\Fref{fig:diversity} (a), we analyze the change in cluster compactness during 2,000 training iterations, comparing $\mathcal{L}_{dm}$ with $\mathcal{L}_{dir}$. During training, we observe that both losses drop the sample diversity of the source domain during the adaptation, indicating that generating diverse samples is challenging in the zero-shot setting. However, we observe that the slope of $\mathcal{L}_{dm}$ is more gentle compared to $\mathcal{L}_{dir}$, verifying that augmenting semantic variations with the original direction indeed alleviates the mode collapse. Moreover, both $\mathcal{L}_{EWC}$ and $\mathcal{L}_{rel}$ are beneficial for mitigating the collapse and sustaining the sample diversity by effectively preserving important source knowledge and relations.

Additional quantitative evaluations on image fidelity metrics, \ie, FID~\cite{heusel2017gans}, precision and recall~\cite{sajjadi2018assessing, simon2019revisiting, kynkaanniemi2019improved}, are included in the  supplementary material.


\noindent\textbf{User study.} To further evaluate the fidelity of generated samples of the target domain, we conduct a user study on the ``Cat-to-Dog'' adaptation scenario with 58 subjects. We present the generated images from ours and competitors to users to select the best one corresponding to the target domain. As a result, $86.76\%$ of participants favored our results as shown in~\Fref{fig:diversity} (b). We also present the participants with 4 images from each method and asked them to choose the one with diverse characteristics. Again, most of the respondents have chosen our method, which indicates that the generated images by ours suffer less from the mode collapse problem and well represent various features of the target domain. The details of the questionnaires for the user study are provided in the supplementary material.

\begin{figure}[t]
  \centering
  \includegraphics[clip=true,width=1.0\linewidth]{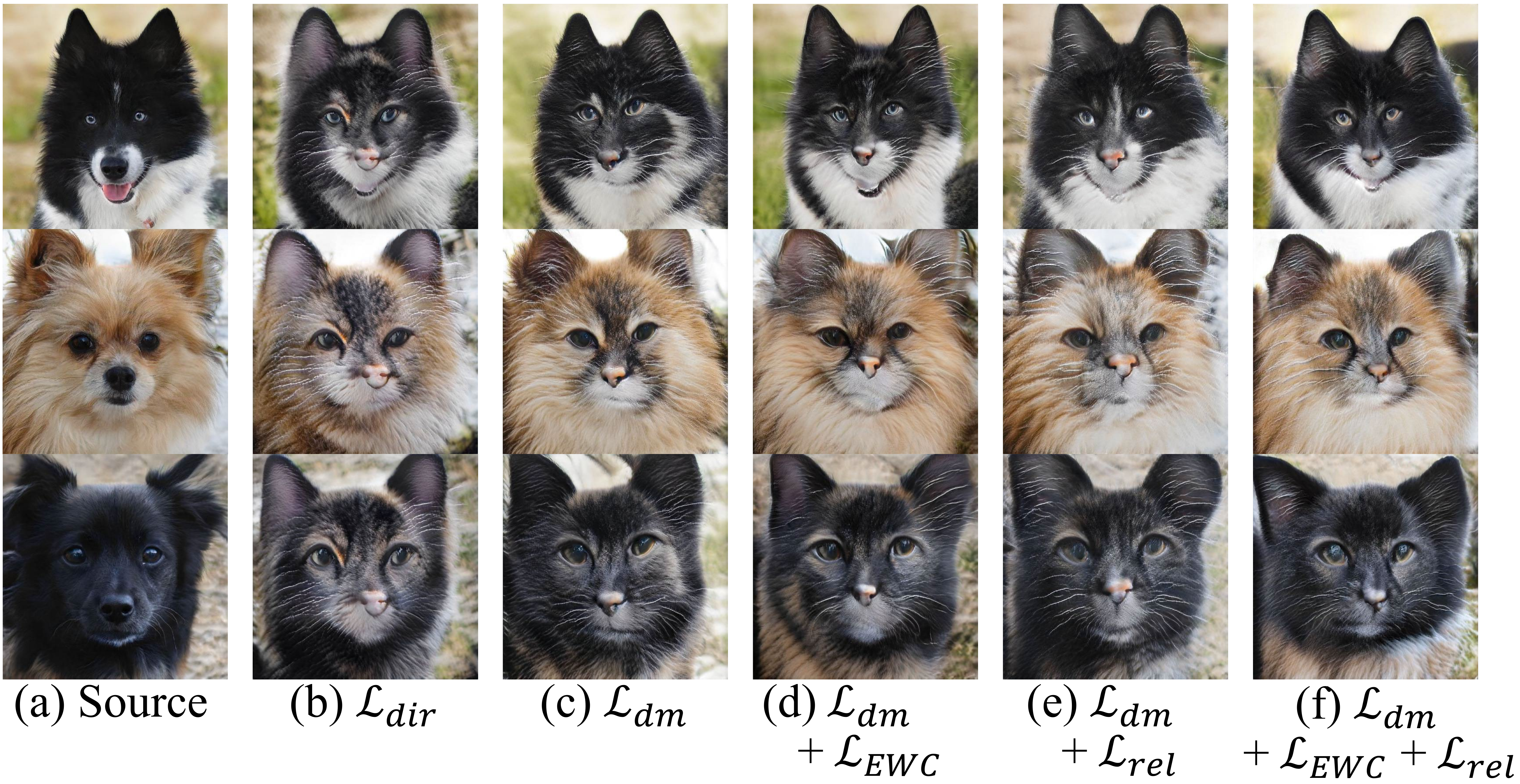}
  \caption{Qualitative ablation study on the ``Dog-to-Cat'' scenario.}
  \label{fig:ablation}
\end{figure}

\begin{figure*}[t]
  \centering
  \includegraphics[clip=true,width=0.98\linewidth]{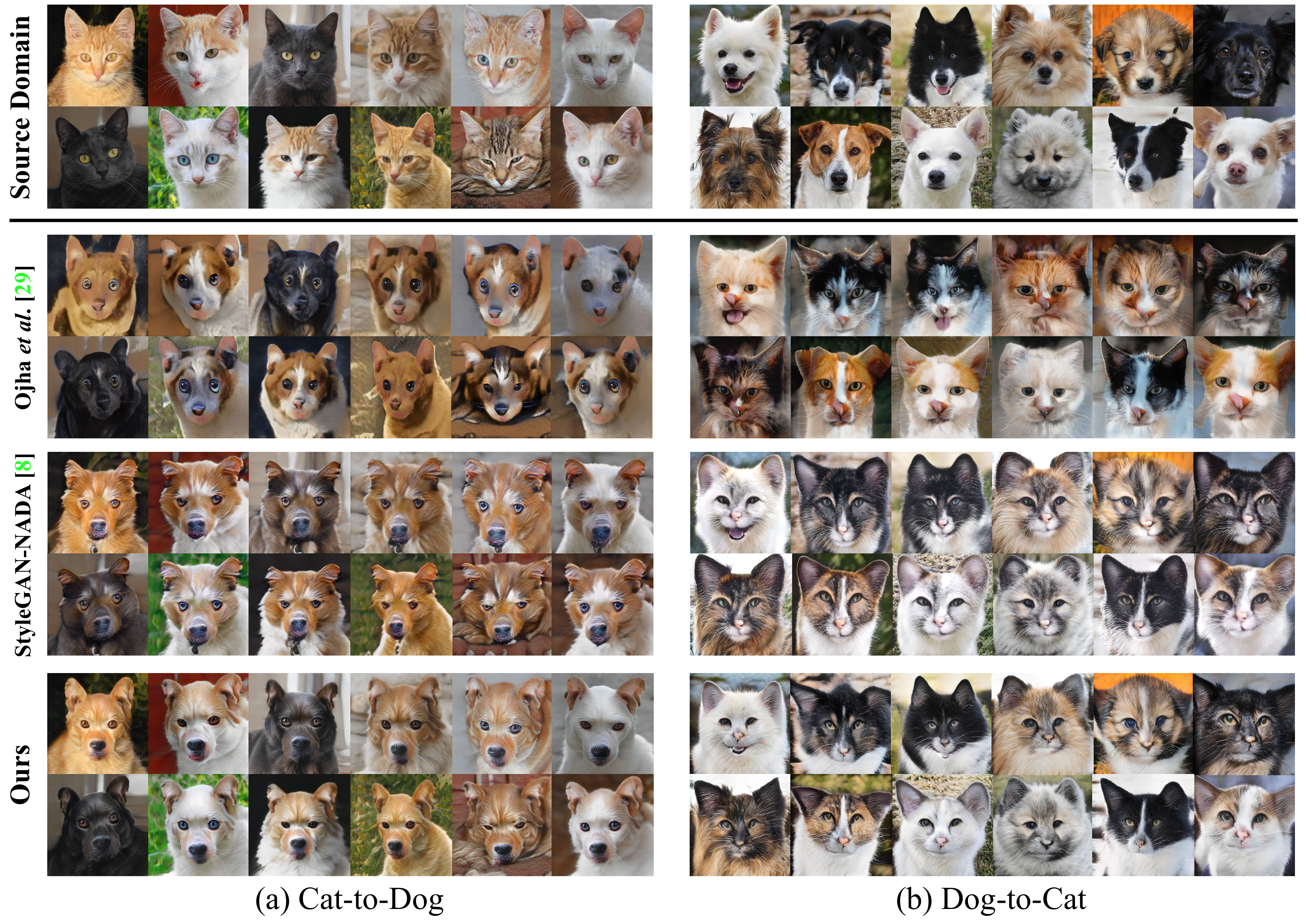}
  \caption{Qualitative comparison on AFHQ datasets~\cite{choi2020stargan}. We compare our model to the previous state-of-the-art zero-shot~\cite{gal2021stylegan} and few-shot~\cite{ojha2021few} methods under the two scenarios: (a) ``Cat-to-Dog'' and (b) ``Dog-to-Cat''. Here Ojha~et~al.~\cite{ojha2021few} is trained in the 10-shot setting.}
  \label{fig:qualitative_comparison}
\end{figure*}

\begin{figure*}[t]
  \centering
  \includegraphics[clip=true,width=0.97\linewidth]{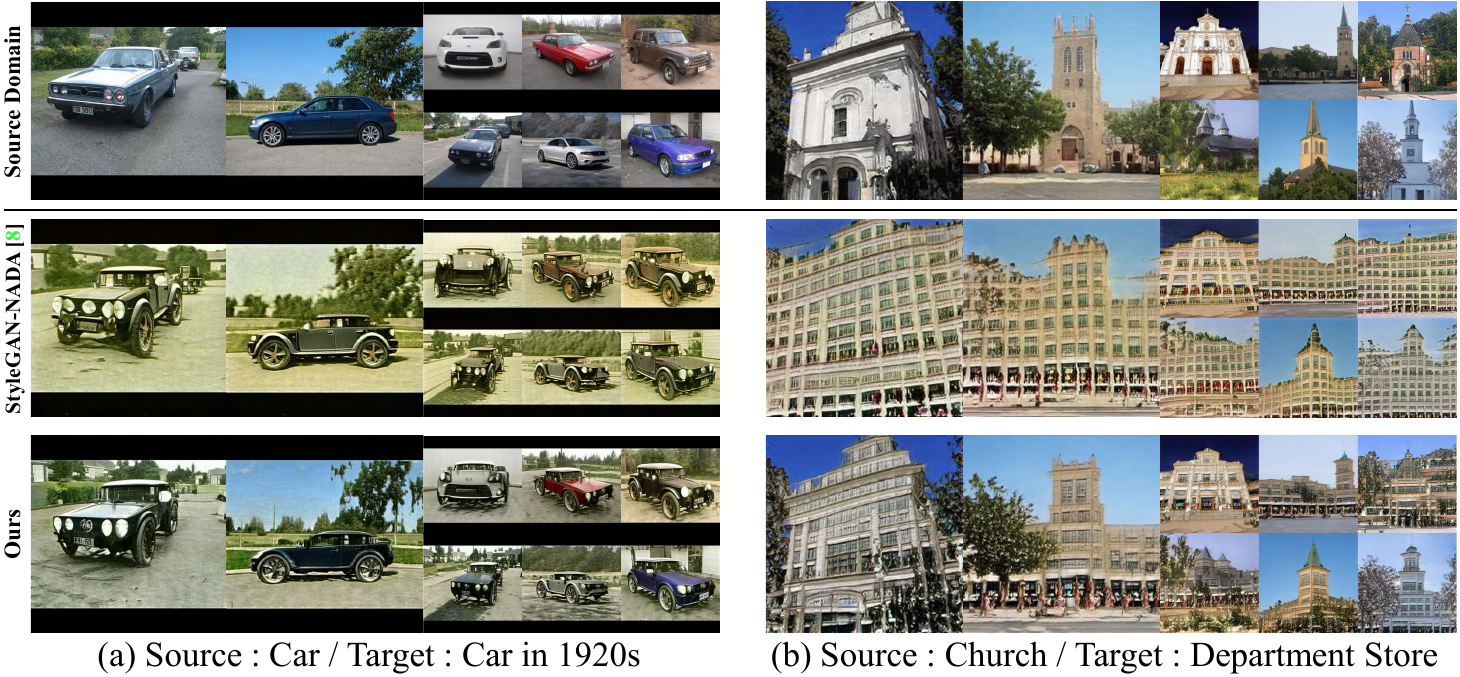}
  \caption{Qualitative comparison in the object translation scenarios. The source generators are pre-trained on LSUN-Car and LSUN-Church datasets~\cite{karras2019style} and adapted to the target text descriptions "Car in 1920s" and "Department Store", respectively. Noticeably, the proposed framework enhances both the diversity and quality of the generated samples.}
  \label{fig:qualitative_comparison_t2}
\end{figure*}
\begin{figure}[t]
  \centering
  \includegraphics[clip=true,width=0.9\linewidth]{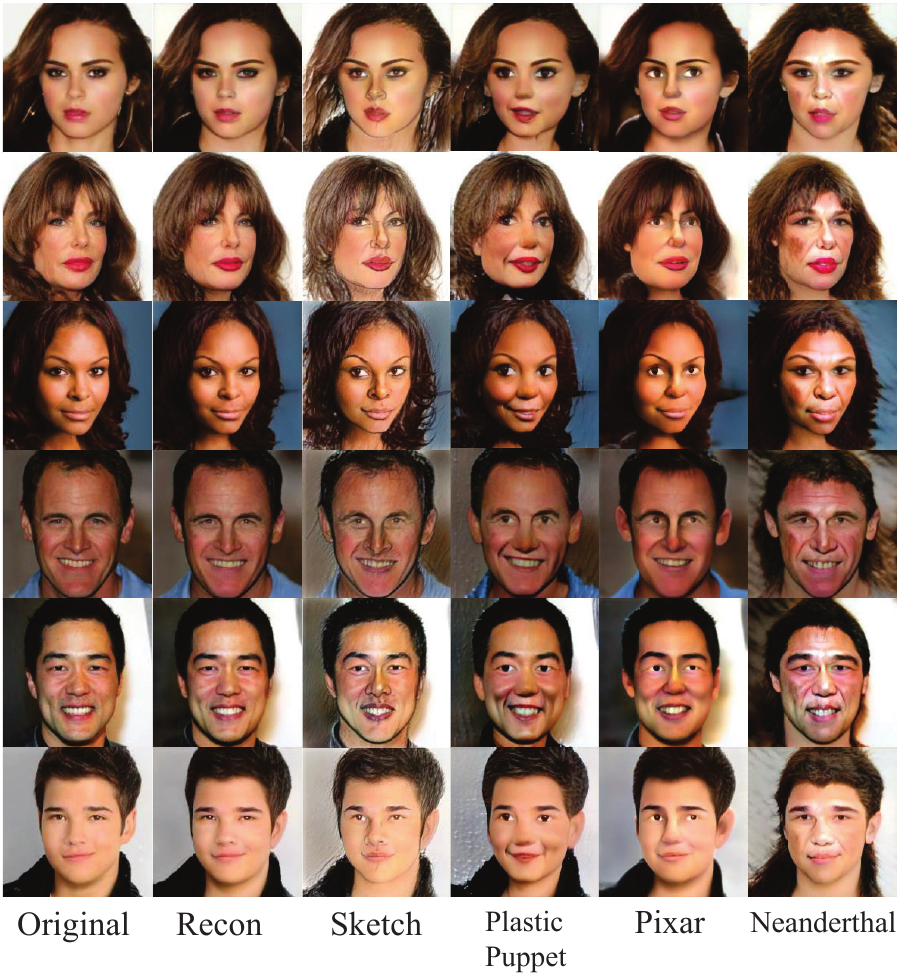}
  \caption{Image manipulation results on different target domains. The real images are sampled from CelebA~\cite{liu2015faceattributes} dataset and inverted into latent space via psp~\cite{richardson2021encoding} trained on FFHQ~\cite{karras2019style}.}
  \label{fig:manipulation}
\end{figure}

\subsection{Ablation Studies.}
We conduct ablation studies on our method with the ``Dog-to-Cat'' scenario qualitatively to verify the effect of each proposed component, whose results are shown in~\Fref{fig:ablation}. When trained with the directional loss $\mathcal{L}_{dir}$, the generator $G_{trg}$ is guided with only a single direction and loses the sample diversity, synthesizing very similar cat faces with common attributes, \eg, purple ears. On the other hand, $\mathcal{L}_{dm}$ successfully mitigates the problem and helps the generator to generate cat faces with different features, \eg, eyes, ears, and facial directions. In addition with $\mathcal{L}_{EWC}$, quality and diversity is improved with the help of important parameters of the source generator which accounts for naturalness. Furthermore, $\mathcal{L}_{rel}$ emphasizes the characteristics of source images, such as eyes and facial appearances. With the all components combined together, our model produces realistic images of the target domain (cat) while preserving diverse diverse attributes of the source domain (dog).

\subsection{Qualitative Results}

In~\Fref{fig:qualitative_comparison}, we compare our method qualitatively with StyleGAN-NADA~\cite{gal2021stylegan} and Ojha~et~al.~\cite{ojha2021few} on ``Cat-to-Dog'' and ``Dog-to-Cat'' scenarios. Note that we randomly sample 10 training images from the AFHQ dataset to train the few-shot GAN adaptation method of Ojha~et~al. We present the generated samples from the same latent codes for fair comparisons. As shown in~\Fref{fig:qualitative_comparison} (a), the generated samples from StyleGAN-NADA largely lose the diversity, showing dog images that resemble each other. Consistently, the generated cats by StyleGAN-NADA in~\Fref{fig:qualitative_comparison} (b) show very similar facial characteristics and expressions without discrimination. Meanwhile, Ojha~et~al. preserve diversity but fail to achieve high quality in both scenarios. In contrast, our method successfully generates realistic images with diverse characteristics of the target domain.

Also in~\Fref{fig:qualitative_comparison_t2}, we display the qualitative results with object adaptation scenarios, \ie, ``Car-to-Car in 1920s'' and ``Church-to-Department Store'', using the source generator trained on LSUN~\cite{yu2015lsun} dataset. Since StyleGAN-NADA heavily depends on the single target text feature, the results lack the diversity while reflecting the common design. For example, the diverse characteristics of the cars in the source domain, \eg, shapes and colors, are diminished after adaptation (\Fref{fig:qualitative_comparison_t2} (a)). In addition, the generated department stores all have same repetitive windows, while the original contexts are collapsed with the entire image filled with the building. In contrast, our model synthesizes more natural images of the target objects with different designs reflecting the source contextual variations. More qualitative results and comparisons in various adaptation scenarios are provided in the supplementary material.

To demonstrate that the proposed framework can also be utilized for text-guided image editing, we show manipulation results of real images with the text prompts. We employ StyleGAN \cite{karras2019style} pretrained on FFHQ as the source generator and sample the images from the CelebA test split~\cite{liu2015faceattributes} as the manipulation target. To embed the real images into the latent space, we exploit GAN inversion methods \cite{richardson2021encoding, alaluf2021restyle, abdal2019image2stylegan}. Afterwards, we feed-forward the obtained latent codes to the target generator adapted to the designated text to get the final results. Since the mapping network and the latent space remain unchanged, the source and target images from the same latent code share the same identity. On the other hand, the parameters of the target generator are updated to align the image editing direction with the text-guided direction from the source to the target domain. As shown in~\Fref{fig:manipulation}, our framework successfully translates the real images into various target domains while preserving the personal characteristics.

\section{Conclusion}
\label{sec:conclusion}
In this paper, we proposed a novel zero-shot GAN adaptation framework that can generate diverse samples of the target domain. Specifically, we introduced a novel method to find semantic variations of the target text in CLIP embedding space and propose a directional moment loss for encouraging the target generator to learn the diverse characteristics of the target domain. Furthermore, in order to preserve the knowledge obtained from the source domain, we employ elastic weight consolidation (EWC) to regularize the drastic parameter updates of the generator. In addition, we introduce a relation consistency loss for more diversity. Through experiments on various adaptation scenarios, we demonstrate that our proposed methods ensure the target sample diversity both qualitatively and quantitatively. In addition, our model achieves a new state-of-the-art on the task of zero-shot GAN adaptation.

\noindent\textbf{Acknowledgements.} 
{\small This research was partly supported by the MSIT (Ministry of Sci-
ence, ICT), Korea, under the High-Potential Individuals Global
Training Program (No. 2021-0-01696) supervised by the IITP (Insti-
tute for Information \& Communications Technology Planning \&
Evaluation), and the National Research Foundation of Korea grant funded by the Korean government (MSIT) (No. 2022R1A2B5B02001467). This project is supported by Microsoft Research Asia.}

{\small
\bibliographystyle{ieee_fullname}
\bibliography{egbib}
}

\clearpage



\appendix


\section{Limitations}

Our proposed directional moment loss $\mathcal{L}_{dm}$ is designed to align the image directions with the text directions.
However, it could be very hard or even infeasible to achieve complete alignment between two modalities in some scenarios with large domain gaps, e.g., ``Dog-to-Joker'' that share few common visual concepts and require significant content changes.
Meanwhile, the adaptation process requires the expert intervention to determine the language description of the source domain as well as appropriate training iterations for the desired result, due to the lack of quality measures.
In addition, since our directional moment loss $\mathcal{L}_{dm}$ solely depends on the guidance provided by CLIP~\cite{radford2021learning} text encoder, there might exist risks of inheriting underlying biases which can cause fairness and privacy issues. Capturing and alleviating the bias for GAN adaptation is an important research topic that is beyond the scope of this work.

\section{Quantitative evaluation.}
\begin{figure*}[h]
  \centering
  \includegraphics[clip=true,width=0.98\linewidth]{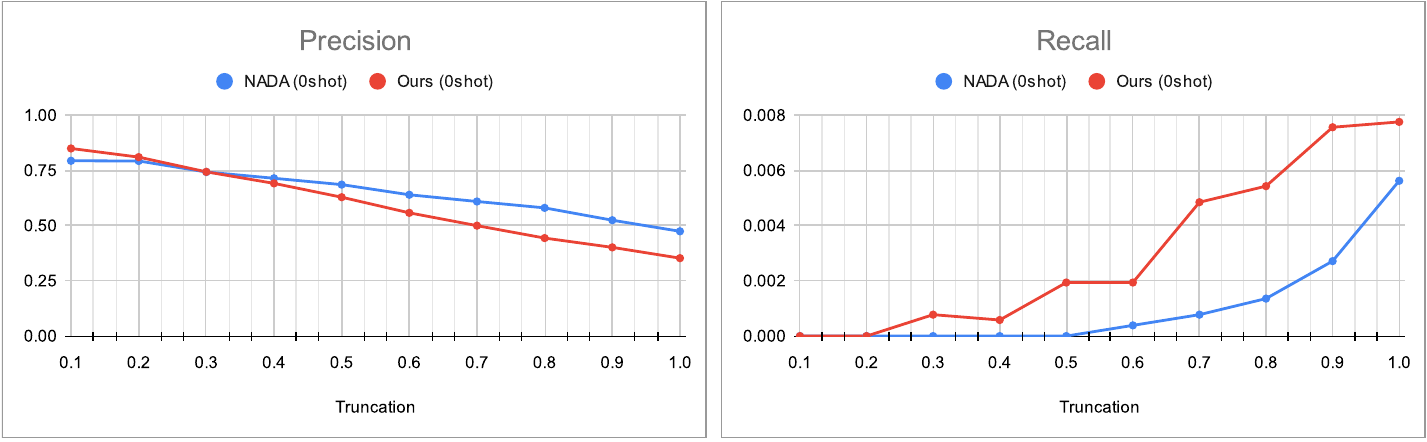}
  \caption{Precision and Recall~\cite{kynkaanniemi2019improved} with varying truncation rates in the ``Dog-to-Cat'' scenario on AFHQ~\cite{choi2020stargan}.}
  \label{fig:pr_curve}
\end{figure*}

For quantitative quality evaluation, we measure Fréchet Inception Distance (FID)~\cite{heusel2017gans} in the ``dog-to-cat'' scenario between the generated samples and the AFHQ-Cat~\cite{choi2020stargan} training split. With 5 independent runs using different random seeds, we generated the same number of images with the real dataset, and then computed the mean and standard deviation of FID. As shown in Table~\ref{tab:fid}, we achieved better FID in zero-shot setting with the help of learned semantic variations as well as source knowledge for diverse characteristics. Moreover, we accomplished the state-of-the-art in the 10-shot setting, demonstrating scalability and superiority. To further evaluate quality and diversity, we respectively plot the precision and recall~\cite{sajjadi2018assessing, simon2019revisiting, kynkaanniemi2019improved} for each truncation rate in Figure~\ref{fig:pr_curve}.  Our framework reports comparable precision with the baseline while showing improvements in recall rate, which is closely related to sample diversity. 

\begin{table}[t]
\caption{FIDs~\cite{heusel2017gans} under the ``Dog-to-Cat'' scenario on AFHQ~\cite{choi2020stargan}.}
  \label{tab:fid}
  \centering
\resizebox{1.0\linewidth}{!}{
\begin{tabular}{l|c}
\toprule
Methods & FID ($\downarrow$)\\
\midrule\midrule
StyleGAN-NADA~\cite{gal2021stylegan} & 70.225$_{\pm\text{0.242}}$\\
Ours & 61.347$_{\pm\text{0.317}}$\\
\midrule
Ojha et al. (10-shot)~\cite{ojha2021few} & 57.196$_{\pm\text{0.249}}$\\
StyleGAN-NADA (10-shot)~\cite{gal2021stylegan} & 60.500$_{\pm\text{0.191}}$\\
Ours (10-shot) & \textbf{45.084}$_{\pm\text{0.108}}$\\
\bottomrule
\end{tabular}
}
\end{table}

\section{Hyperparameter ablations.} 
The quantitative analyses on hyperparameters are provided in Table~\ref{tab:hyperparameter}. Note that $\lambda_{EWC}$ affects the level of source characteristic preservation, while $\lambda_{rel}$ is related to the degree of maintained semantic relationship between images before and after adaptation. 
We meticulously set these weighting factors to guarantee that all losses ($\mathcal{L}_{dm}$, $\mathcal{L}_{EWC}$, $\mathcal{L}_{rel}$) are balanced in terms of their magnitude. In addition, enlarging the number of semantic variations $K$ showed increased fidelity with the excavated semantics of the target domain. However, setting a large $K$ makes the semantic variation learning more challenging, resulting in a slightly quality decrease. Meanwhile, the perturbation strength $\epsilon$ is an important parameter for optimizing semantic variation.
A large $\epsilon$ makes $\mathcal{L}_{cons}$ optimization difficult, while a small $\epsilon$ causes fast convergence and minimal semantic differences after perturbation.
To discover meaningful semantic variations, we empirically set the value $\epsilon$ to $||E_{T}(t_{trg})||_{2}$, which has been shown to enable sufficient convergence within 2,000 iterations.

\begin{table}[t]
\caption{FIDs~\cite{heusel2017gans} under the ``Dog-to-Cat'' scenario on AFHQ~\cite{choi2020stargan}.}
  \label{tab:hyperparameter}
  \centering
\resizebox{0.95\columnwidth}{!}{%
\begin{tabular}{llllll}
\hline
\multicolumn{6}{c}{FID ($\downarrow$)}                                                                                                                                                \\ \hline
\multicolumn{2}{l|}{$\lambda_{EWC}$}                              & \multicolumn{2}{l|}{$\lambda_{rel}$}                             & \multicolumn{2}{l}{$K$ (number of $z^{i}$)} \\ \hline
\multicolumn{1}{l|}{$10^5$}        & \multicolumn{1}{l|}{94.317} & \multicolumn{1}{l|}{$1$}          & \multicolumn{1}{l|}{82.900} & \multicolumn{1}{l|}{2}          & 81.840  \\
\multicolumn{1}{l|}{$10^6$}        & \multicolumn{1}{l|}{77.898} & \multicolumn{1}{l|}{$10$}         & \multicolumn{1}{l|}{76.590} & \multicolumn{1}{l|}{4}          & 71.793  \\
\multicolumn{1}{l|}{$10^7$ (Ours)} & \multicolumn{1}{l|}{\textbf{61.347}} & \multicolumn{1}{l|}{$100$ (Ours)} & \multicolumn{1}{l|}{\textbf{61.347}} & \multicolumn{1}{l|}{6 (Ours)}   & \textbf{61.347}  \\
\multicolumn{1}{l|}{$10^8$}        & \multicolumn{1}{l|}{81.758} & \multicolumn{1}{l|}{$1000$}       & \multicolumn{1}{l|}{81.758} & \multicolumn{1}{l|}{8}          & 66.698 \\ \hline
\end{tabular}%
}
\end{table}

\section{Few-shot GAN adaptation.} 
To further verify the effectiveness, we extended our framework to few-shot GAN adaptation, \ie, 10-shot, 5-shot and 1-shot settings. For each target image, we extract feature from the image encoder and conduct semantic variation learning process. 
As reported in Table~\ref{tab:fid}, our framework achieved FID score of 45.084 in 10-shot setting, considerably surpassing both StyleGAN-NADA~\cite{gal2021stylegan} and the Ojha~et~al.~\cite{ojha2021few}. It is also notable that we scored 61.347 FID in zero-shot setting comparable to few-shot methods.

\begin{figure}[h]
  \centering
  \includegraphics[clip=true,width=1.0\linewidth]{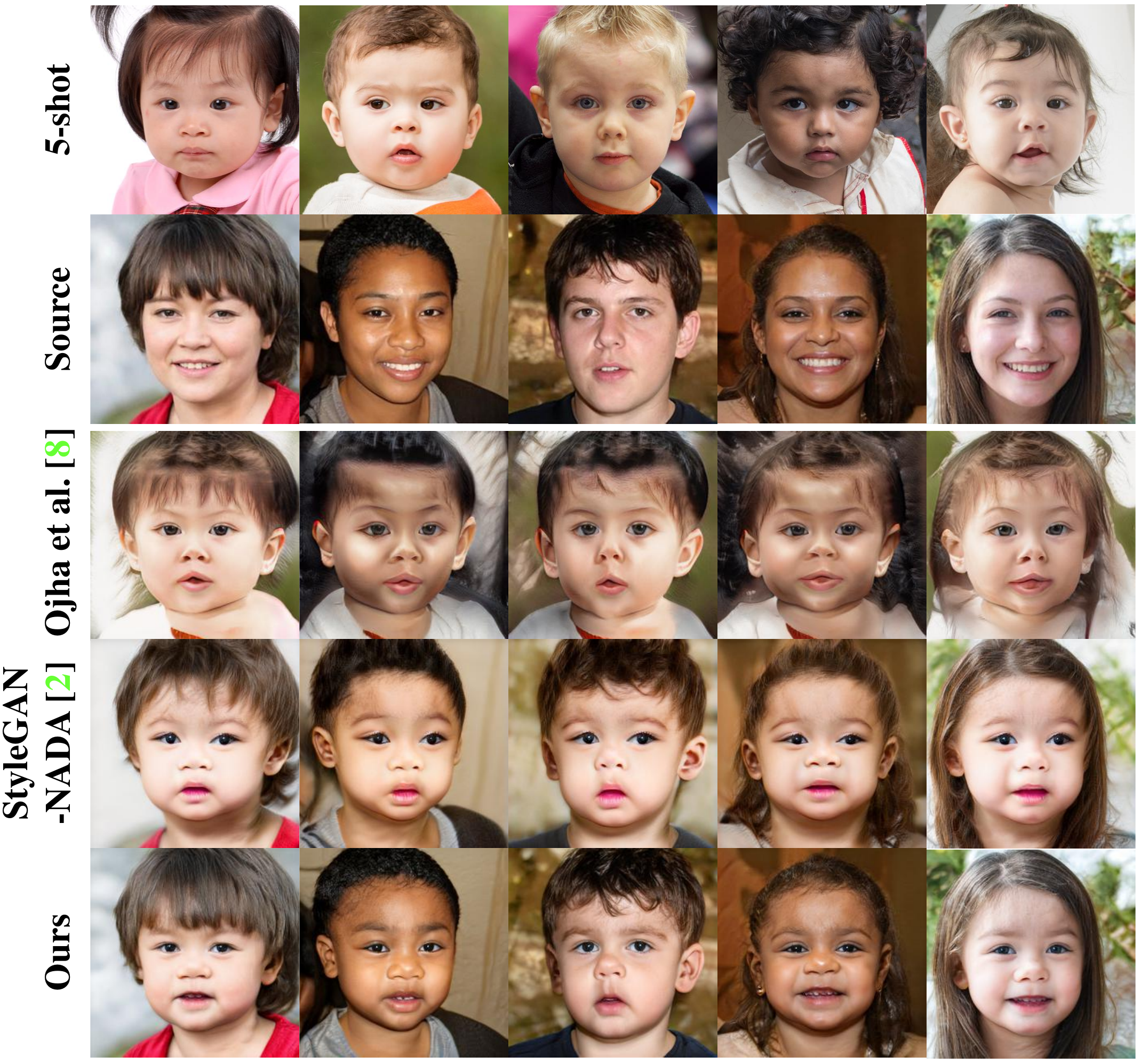}
  \caption{Results on 5-shot GAN adaptation scenario with the StyleGAN2 trained on FFHQ.}
  \label{fig:5_shot}
\end{figure}

\begin{figure}[h]
  \centering
  \includegraphics[clip=true,width=1.0\linewidth]{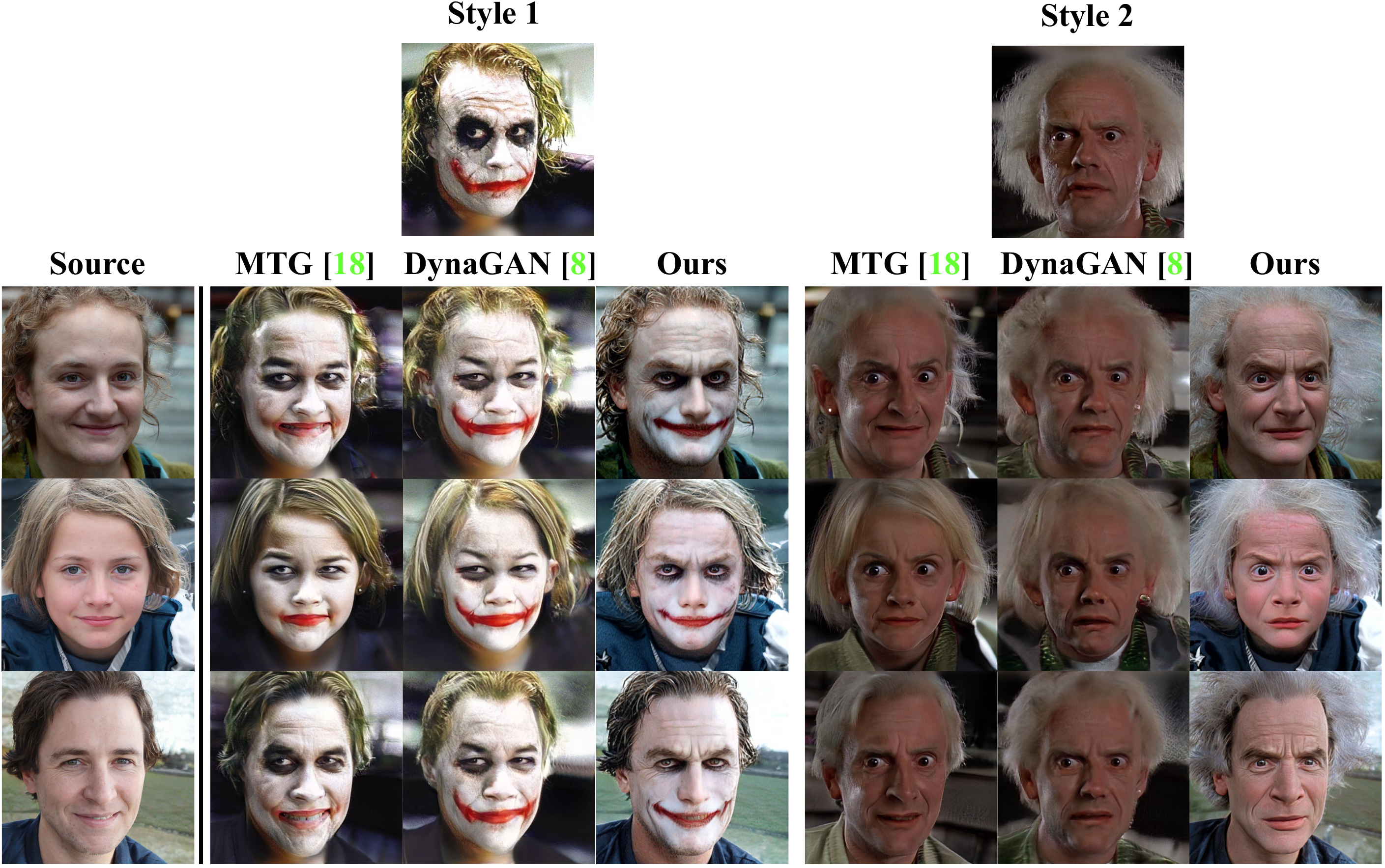}
  \caption{Results on 1-shot GAN adaptation scenarios with the StyleGAN2 trained on FFHQ.}
  \label{fig:1_shot}
\end{figure}

Moreover, in 5-shot and 1-shot settings, we present qualitative results respectively in Figure~\ref{fig:5_shot} and~\ref{fig:1_shot}. Remarkably, our method shows more visually favorable results with diverse facial characteristics, e.g., emotional expressions, compared to previous methods~\cite{ojha2021few, gal2021stylegan} as shown in Figure~\ref{fig:5_shot}. Also compared to MTG~\cite{zhu2021mind} and DynaGAN~\cite{kim2022dynagan} in 1-shot setting (Figure~\ref{fig:1_shot}), it is notable that our method has strength in both utilizing diverse semantic information of source contents and maintaining original identity, while still being able to capture the style of target samples (\eg, face of the joker, hairstyle of the doctor brown).

\section{Additional Qualitative Results}
\begin{figure*}[ht]
  \centering
  \includegraphics[clip=true,width=1.0\linewidth]{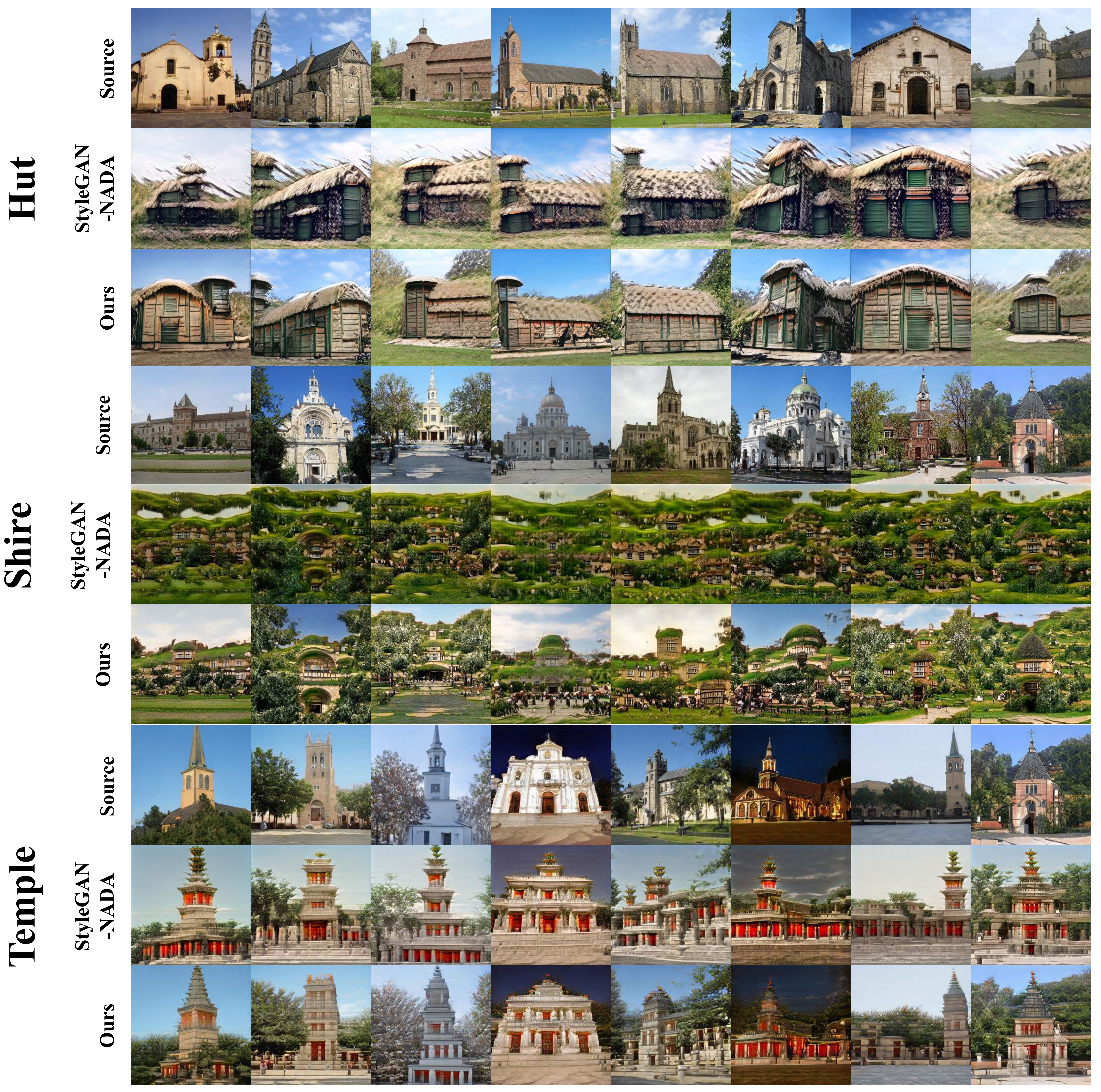}
  \caption{Qualitative results in the adaptation scenarios of from the source generator trained on LSUN-Church dataset~\cite{yu2015lsun} to different buildings, i.e, ``Hut'' and ``Temple'', and to fictitious village of ``Shire''. The results of StyleGAN-NADA~\cite{gal2021stylegan} are sharing a specific design or global characteristic of the target domain, e.g., green doors and walls of huts. On the other hand, the proposed framework synthesizes buildings with more diverse textural details. Moreover, various contexts of the source domain are well inherited and fully utilized to generate satisfactory results.}
  \label{fig:suppl_lsun}
\end{figure*}

\begin{figure*}[ht]
  \centering
  \includegraphics[clip=true,width=1.0\linewidth]{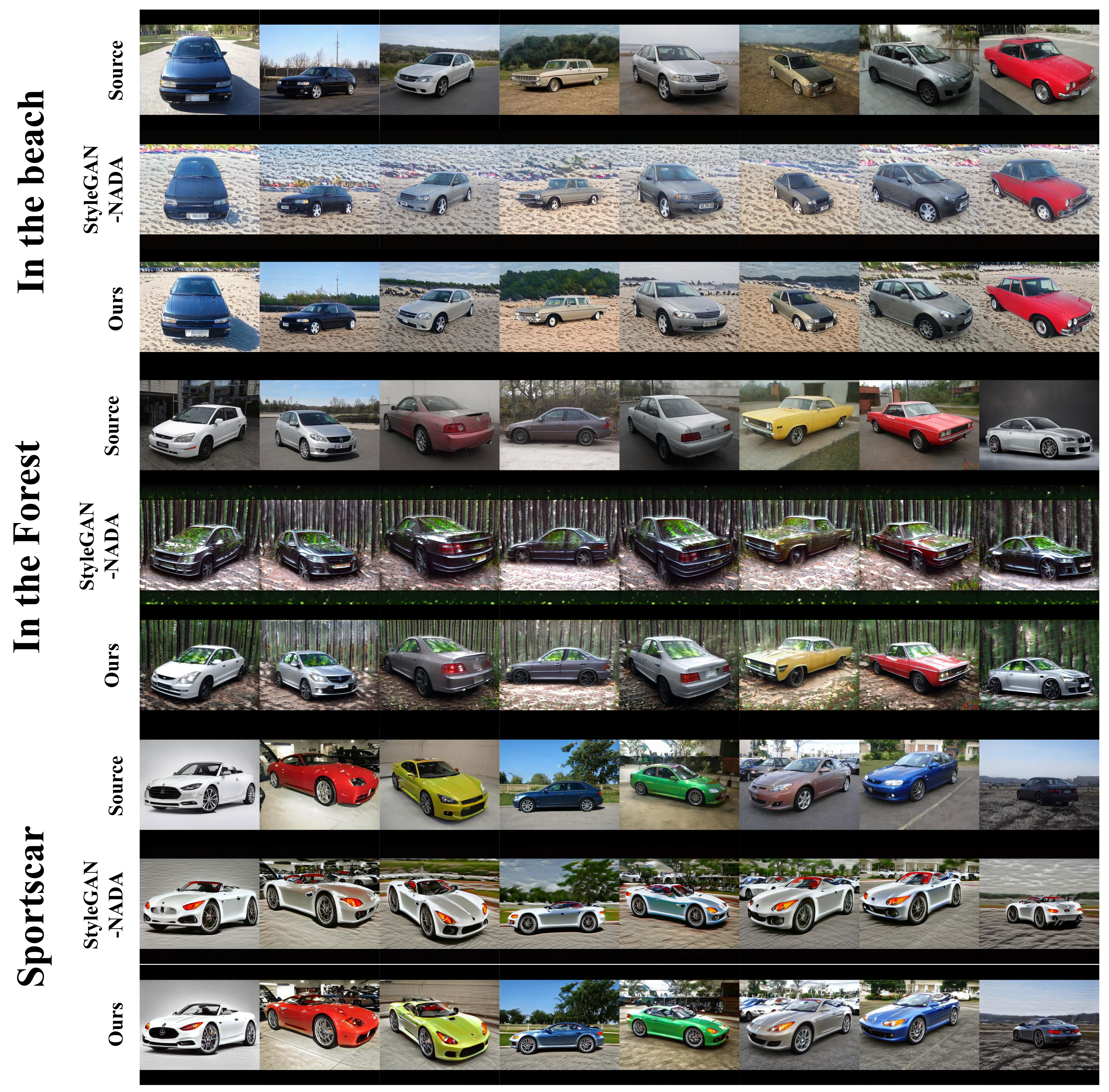}
  \caption{Qualitative results in the adaptation scenarios of from the source generator trained on LSUN-Car dataset~\cite{yu2015lsun} to different backgrounds, i.e, ``Car in the beach'' and ``Car in the forest'', and appearance translation to ``Sportscar''. StyleGAN-NADA~\cite{gal2021stylegan} samples fails in succeeding context of the source domain, filling the overall image even in the sky region. Also in ``Car-to-Car in the forest'' and ``Car-to-Sportscar'', the underlying bias of the target text prompt are reflected to StyleGAN-NADA, e.g., dark car in the forest and same the sportscar design. In contrary, our framework generates results with more diverse designs and characteristics.}
  \label{fig:suppl_car}
\end{figure*}

\begin{figure*}[ht]
  \centering
  \includegraphics[clip=true,width=1.0\linewidth]{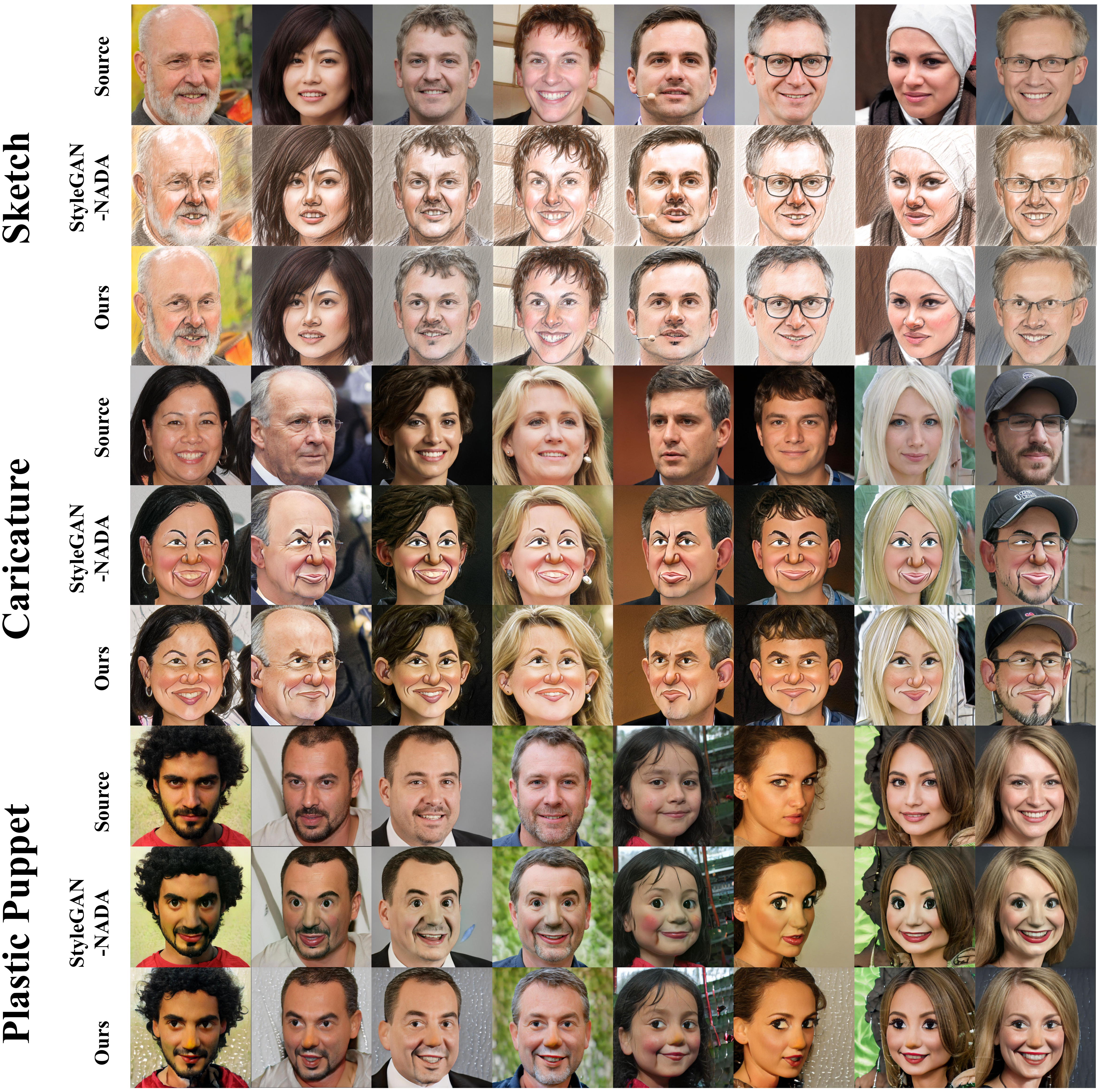}
  \caption{Generated samples from StyleGAN2~\cite{karras2020analyzing} trained with FFHQ~\cite{karras2019style} in various adaptation scenarios.}
  \label{fig:suppl_ffhq1}
\end{figure*}

\begin{figure*}[ht]
  \centering
  \includegraphics[clip=true,width=1.0\linewidth]{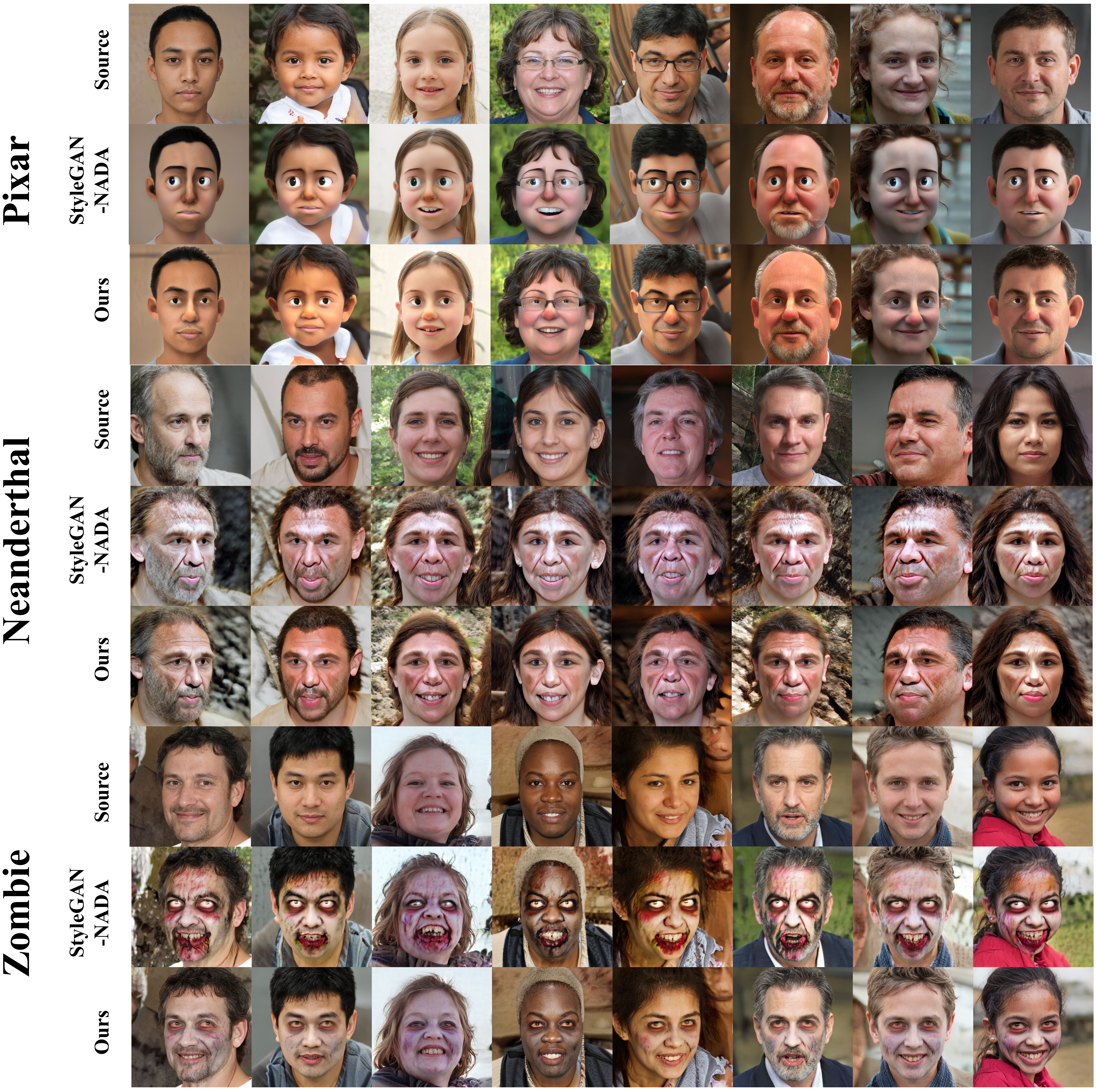}
  \caption{Generated samples from StyleGAN2~\cite{karras2020analyzing} trained with FFHQ~\cite{karras2019style} in various adaptation scenarios.}
  \label{fig:suppl_ffhq2}
\end{figure*}

\begin{figure*}[ht]
  \centering
  \includegraphics[clip=true,width=1.0\linewidth]{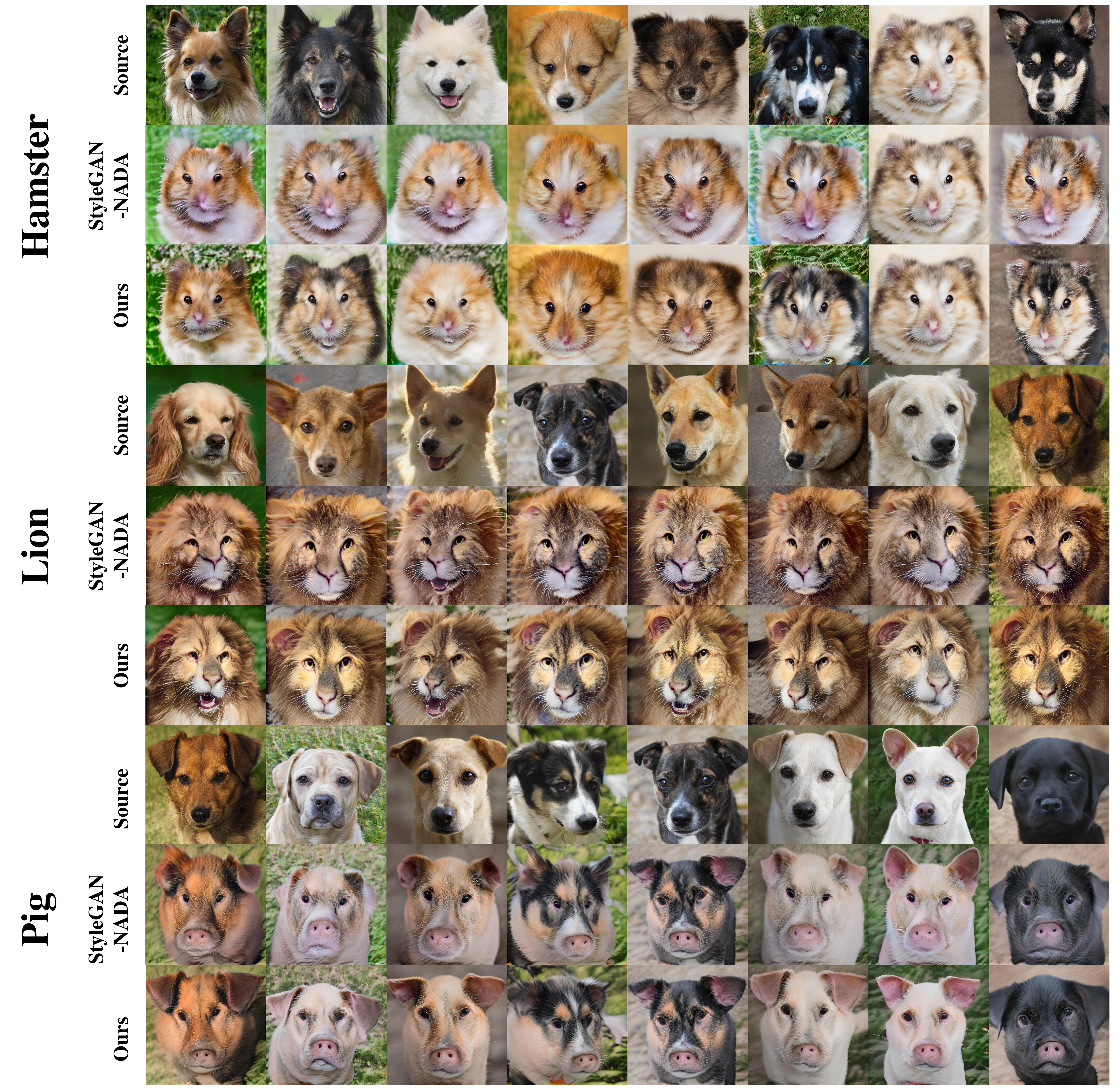}
  \caption{Generated samples from StyleGAN2~\cite{karras2020analyzing} trained with AFHQ-Dog~\cite{choi2020stargan} adapted to various target domains.}
  \label{fig:suppl_afhq1}
\end{figure*}

\begin{figure*}[ht]
  \centering
  \includegraphics[clip=true,width=1.0\linewidth]{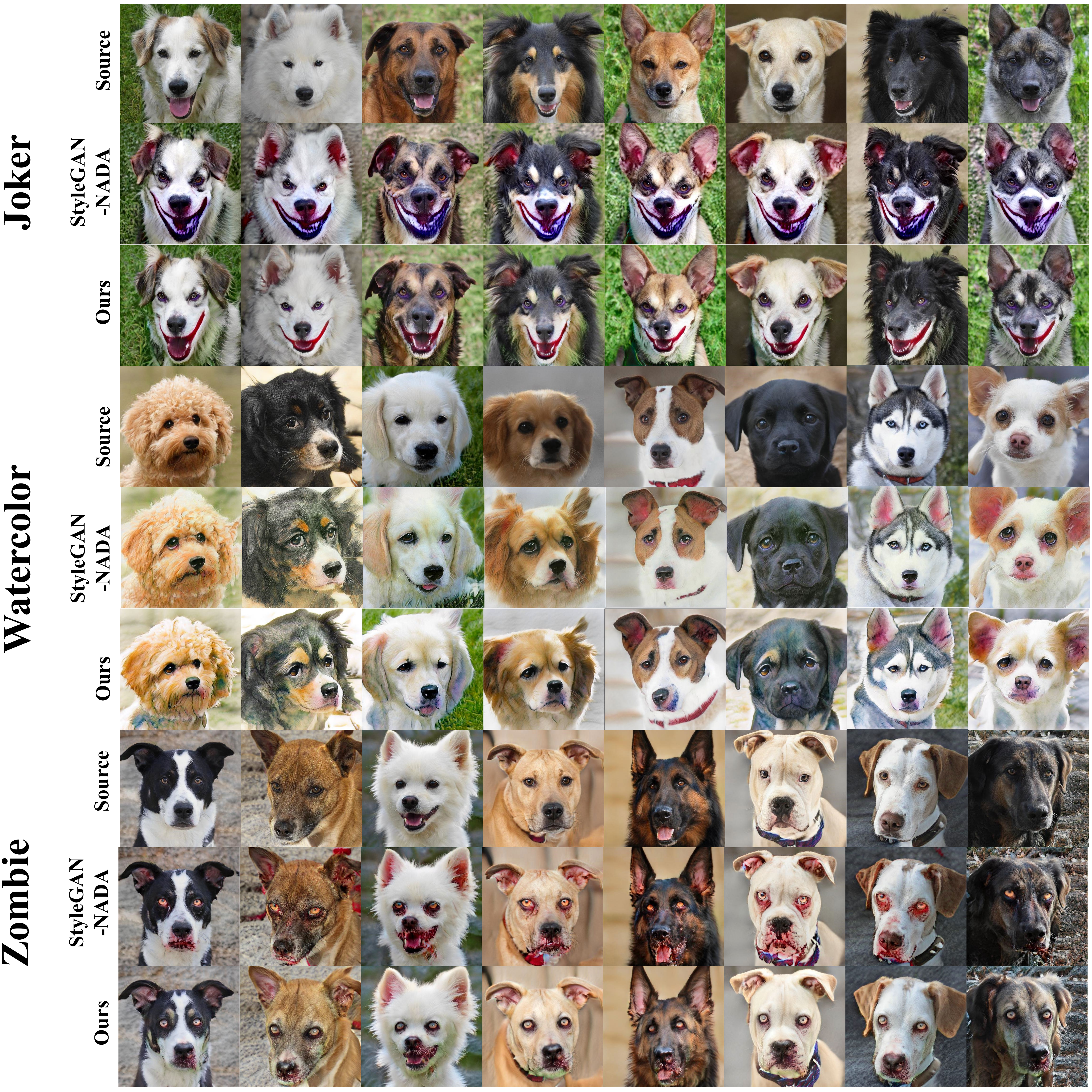}
  \caption{Generated samples from StyleGAN2~\cite{karras2020analyzing} trained with AFHQ-Dog~\cite{choi2020stargan} adapted to various target domains.}
  \label{fig:suppl_afhq2}
\end{figure*}

In this section, we provide further qualitative results in various GAN adaptation scenarios which are not included in the main manuscript in Figures~\ref{fig:suppl_lsun} through \ref{fig:suppl_afhq2}. Most of the hyperparameters are kept unchanged except for $\lambda_{ewc}$, which is adjusted from $10^7$ to $10^6$ for artistic texture manipulation, e.g., ``Photo-to-Sketch'' and ``Photo-to-Caricature''. We can observe that structural layouts are reflected in the results of StyleGAN-NADA and ours. Noticeably, the proposed framework synthesizes buildings with diverse characteristics with the help of $\mathcal{L}_{dm}$ which guides the model with explored semantic variations. Also, the contexts are well preserved by inheriting useful knowledge from the source generator via $\mathcal{L}_{EWC}$ and $\mathcal{L}_{rel}$.

\begin{table}[h]
\caption{Sources and licenses of the utilized models and datasets}
  \label{tab:license}
  \centering
\begin{tabular}{|lc|}
\hline
Models                                                       & License                    \\ \hline
\multicolumn{1}{|l|}{StyleGAN2~\cite{karras2020analyzing}}                             & Nvidia Source Code License \\ \hline
\multicolumn{1}{|l|}{scikit-learn-extra~\cite{varoquaux2015scikit}}                    & BSD 3-Clause               \\ \hline
\multicolumn{1}{|l|}{CLIP~\cite{radford2021learning}}                                  & MIT License                \\ \hline
\multicolumn{1}{|l|}{StyleGAN2-pytorch~\cite{stylegan2-pytorch}}                     & MIT License                \\ \hline
\multicolumn{1}{|l|}{StyleGAN-ADA~\cite{karras2020training}}                          & Nvidia Source Code License \\ \hline
\multicolumn{1}{|l|}{psp~\cite{richardson2021encoding}}                     & MIT License                \\ \hline
\multicolumn{1}{|l|}{Ojha \textit{et al}.~\cite{ojha2021few}} & Adobe Research License     \\ \hline
\multicolumn{1}{|l|}{StyleGAN-NADA~\cite{gal2021stylegan}}                         & MIT License                \\ \hline
\multicolumn{2}{|l|}{Datasets}                                                                                     \\ \hline
\multicolumn{1}{|l|}{FFHQ~\cite{karras2019style}}                                  & CC BY-NC-SA 4.0            \\ \hline
\multicolumn{1}{|l|}{AFHQ~\cite{choi2020stargan}}                                  & CC BY NC 4.0               \\ \hline
\multicolumn{1}{|l|}{LSUN~\cite{yu2015lsun}}                                  & No License               \\ \hline
\multicolumn{1}{|l|}{CelebA~\cite{liu2015faceattributes}}                                  & CC BY-NC-SA 4.0               \\ \hline
\end{tabular}
\end{table}

\section{Questionnaire for User Study}

\begin{figure*}[t]
  \centering
  \includegraphics[clip=true,width=1.0\linewidth]{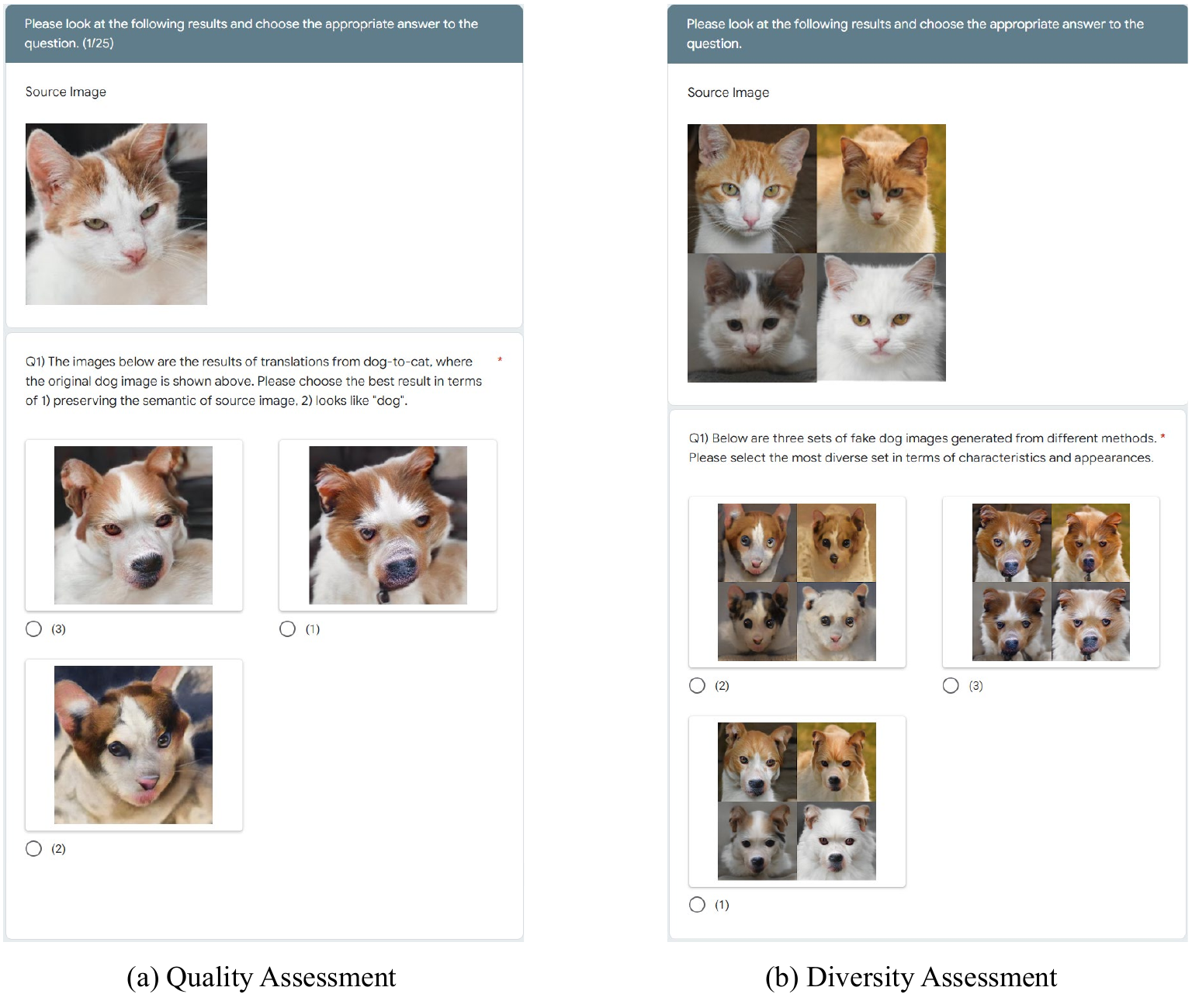}
  \caption{Screenshots of user study on (a) quality assessment and (b) diversity assessment.}
  \label{fig:survey}
\end{figure*}

In~\Fref{fig:survey}, we present the screenshots of questionnaires for the user study. The questionnaire is composed of 25 questions for quality evaluation and 6 questions for diversity assessment. Each question contains the generated samples from Ojha~et~al~\cite{ojha2021few}, StyleGAN-NADA~\cite{gal2021stylegan}, and ours in the ``Cat-to-Dog'' scenario from the same latent codes. To evaluate the quality, we requested users to select the best result that looks most like a ``Dog'' while preserving the content information of the source image~(\Fref{fig:survey}~(a)). For diversity assessment, we provided the participants with a set of 4 images from each method and asked to pick the one showing the most diverse characteristics of dogs~(\Fref{fig:survey}~(b)). Note that we manually shuffled the order of methods for the reliability of the survey.

\section{License}

In Table 1, we specify the source and licenses of the models and datasets used in our work. Note that the FFHQ~\cite{karras2019style} dataset consists of facial images collected from Flickr, which are under permissive licenses for non-commercial purposes. 

\clearpage

\clearpage

\end{document}